\documentclass{article}

\usepackage{microtype}
\usepackage{graphicx}
\usepackage{subfigure}
\usepackage{booktabs} 
\usepackage{listings}
\usepackage{amsmath}
\usepackage{amsthm}
\usepackage{amssymb}
\usepackage{siunitx}
\usepackage{afterpage}
\usepackage{balance}

\usepackage[title]{appendix}

\theoremstyle{definition}
\newtheorem{example}{Example}[section]
\newtheorem{definition}{Definition}[section]

\usepackage{xcolor}
\definecolor{blueish}{RGB}{103, 135, 176}
\definecolor{reddish}{RGB}{ 205, 102, 7}
\usepackage{hyperref}
\hypersetup{colorlinks, linkcolor={reddish!70!black},
	citecolor={blueish!70!black}, urlcolor=black,}

\usepackage[nameinlink,noabbrev]{cleveref}
\crefname{equation}{}{}
\Crefname{equation}{}{}

\usepackage{enumitem}
\makeatletter
\def\namedlabel#1#2{\begingroup #2%
    \def\@currentlabel{#2}%
    \phantomsection\label{#1}\endgroup}
\makeatother

\usepackage[accepted]{icml2020}

\begin{document}

\twocolumn[\icmltitle{TriFinger: An Open-Source Robot for Learning Dexterity}

		
		
	\icmlsetsymbol{equal}{*}
		
	\begin{icmlauthorlist}

		\icmlauthor{Manuel W\"uthrich}{mpi}
		\icmlauthor{Felix Widmaier}{mpi}
		\icmlauthor{Felix Grimminger}{mpi}
		\icmlauthor{Joel Akpo}{mpi}
		\icmlauthor{Shruti Joshi}{mpi}
		\icmlauthor{Vaibhav Agrawal}{mpi}
		\icmlauthor{Bilal Hammoud}{nyu,mpi}
		\icmlauthor{Majid Khadiv}{mpi}
		\icmlauthor{Miroslav Bogdanovic}{mpi}
		\icmlauthor{Vincent Berenz}{mpi}
		\icmlauthor{Julian Viereck}{nyu,mpi}
		\icmlauthor{Maximilien Naveau}{mpi}
		\icmlauthor{Ludovic Righetti}{nyu,mpi}
		\icmlauthor{Bernhard Sch\"olkopf}{mpi}
		\icmlauthor{Stefan Bauer}{mpi}
	\end{icmlauthorlist}

	\icmlaffiliation{mpi}{Max Planck Institute for Intelligent Systems, T\"ubingen, Germany}
	\icmlaffiliation{nyu}{Tandon  School  of  Engineering,  New  York  University,  Brooklyn,  USA}
			
	\icmlcorrespondingauthor{Manuel W\"uthrich}{manuel.wuthrich@gmail.com}

			
\vskip 0.3in] 



\printAffiliationsAndNotice{} 

\begin{abstract}
	Dexterous object manipulation remains an open problem in robotics, despite
	the rapid progress in machine learning during the past decade. We argue that
	a hindrance is the high cost of experimentation
	on real systems, in terms of both time and money. We address this problem by
	proposing an open-source robotic platform which can safely operate without
	human supervision. The hardware is inexpensive (about \SI{5000}[\$]{}) yet
	highly dynamic, robust, and capable of complex interaction with external
	objects. The software operates at 1-kilohertz and performs safety
	checks to prevent the hardware from breaking. The easy-to-use front-end (in
	C++ and Python) is suitable for real-time control as well as deep
	reinforcement learning. In addition, the software framework is largely robot-agnostic and
	can hence be used independently of the hardware proposed herein. Finally, we
	illustrate the potential of the proposed platform through a number of
	experiments, including real-time optimal control, deep reinforcement
	learning from scratch, throwing, and writing.
\end{abstract}

\begin{figure}[h!]
	\centering
	\subfigure[]{
		\includegraphics[width=0.88\columnwidth]{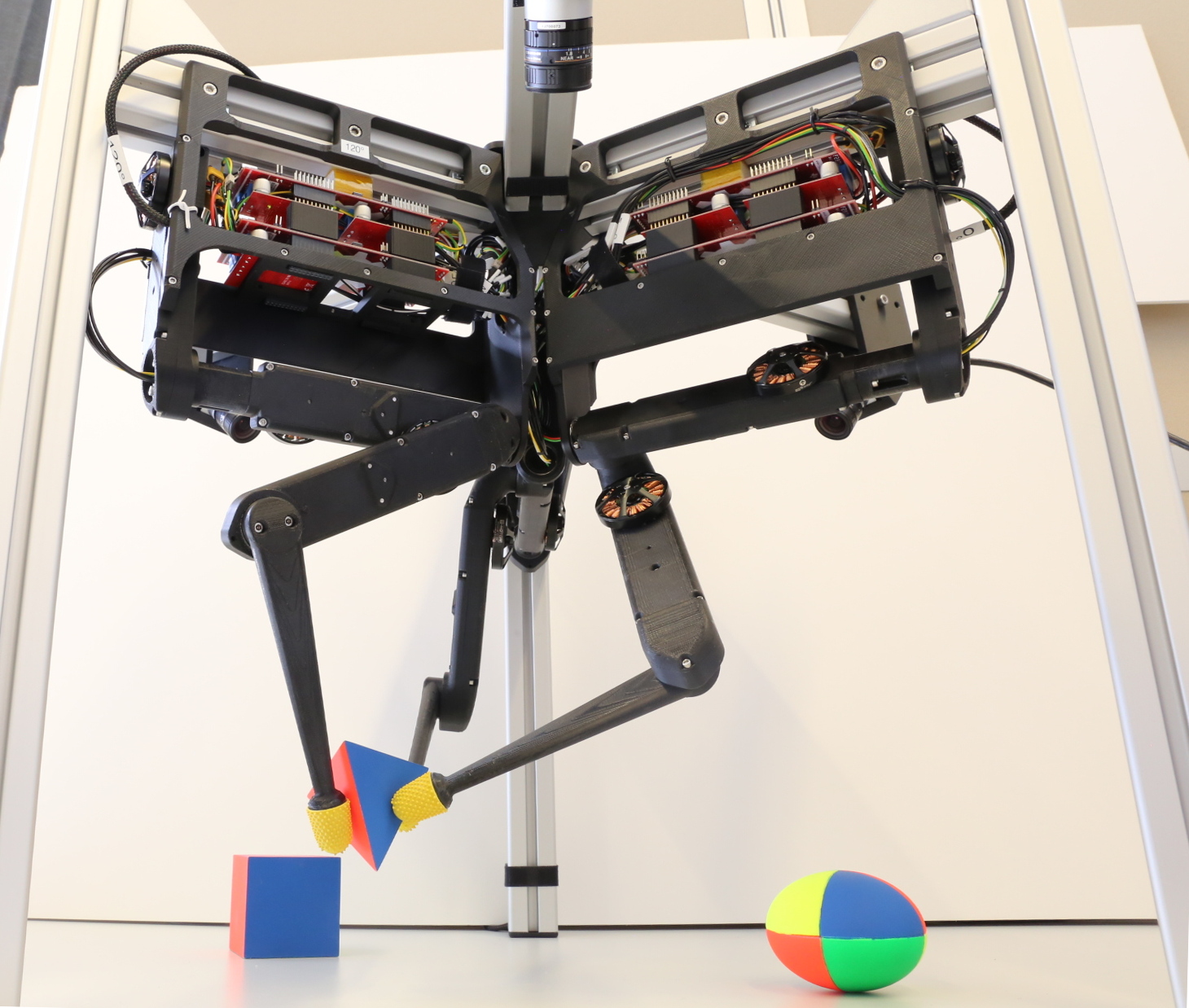}
		\label{fig:manipulation} 
		}\\
	\subfigure[]{
		\includegraphics[width=0.88\columnwidth]{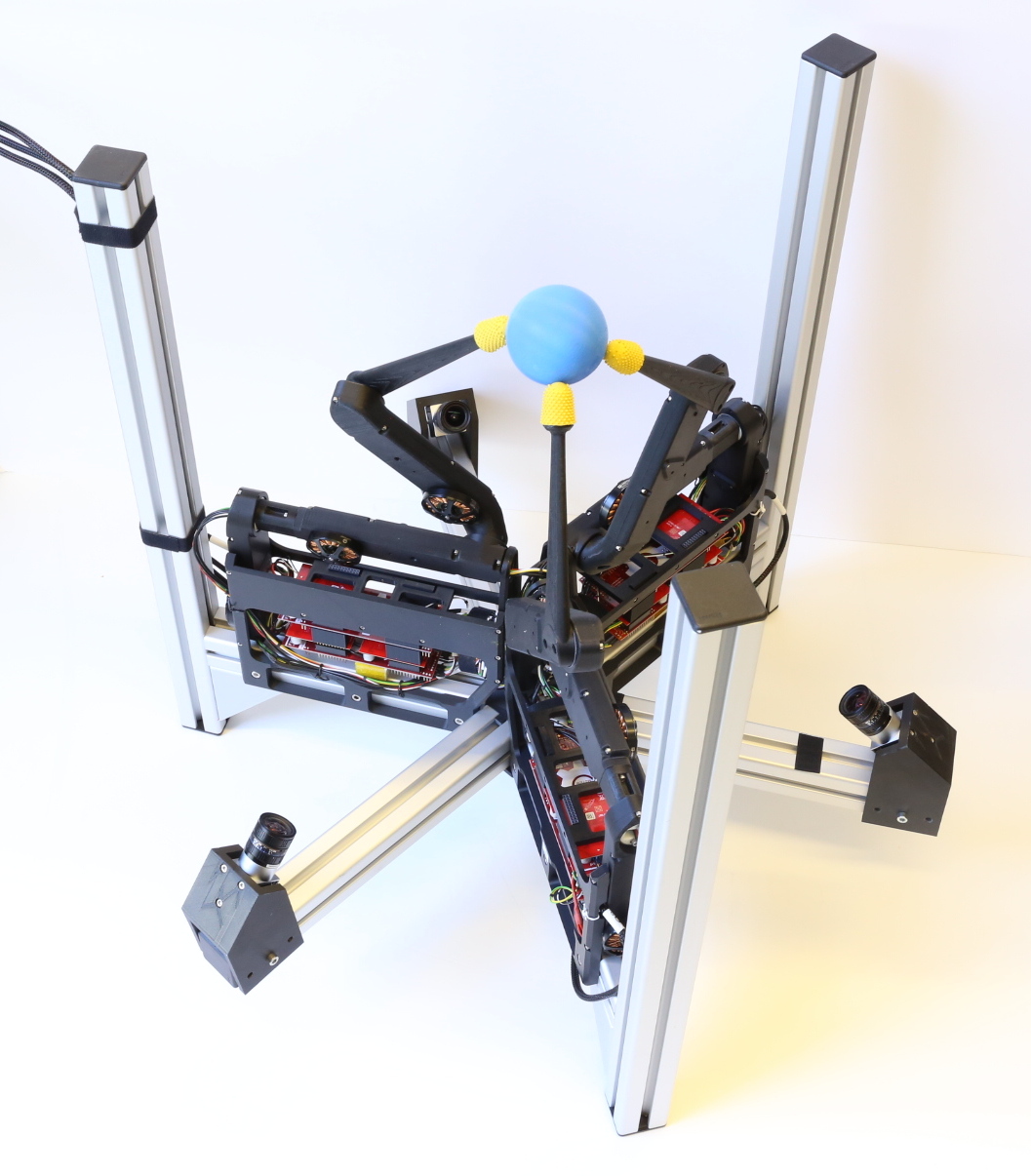}
		\label{fig:catching} 
	}
	\label{fig:overview}
	\caption{(a) The proposed platform can be used for object manipulation on the table. (b) The same platform can be flipped and used for e.g. catching and throwing.}
\end{figure}

\section{Introduction}
In the past decades, researchers have devoted great efforts to making computers
more autonomous, such that they could take care of more and more tasks in our
society. Today, algorithms make a great number of decisions which were made by
humans previously: They decide on the best way to get from one place to another,
what movie we may want to watch next and what articles we might be interested
in. They take strategic decisions for companies, place advertisements and invest
large sums of money. They play and win against us in video games
\cite{mnih2013playing}, chess \cite{campbell2002deep}, and Go
\cite{silver2017mastering}. Yet, the way we construct buildings, clean, cook,
dispose of our trash, plant and harvest food, search and rescue and assist
people with physical disabilities has remained largely unaffected. Where does
this striking contrast come from? 

A difference which catches the eye is that the first class of tasks is primarily
concerned with manipulation of information, whereas the second class is
concerned with manipulation of matter, achieved through physical contact
between the agent and the external world. We believe that a key issue which has
hindered progress in the second class of problems is the high cost of
experimentation. Robotic manipulators are typically very expensive and can break
easily when entering into contact with the external world. Furthermore, they are
usually operated only in the presence of human supervisors, ready to hit the
emergency stop in case something goes wrong. These factors have largely
prohibited systematic large scale experimentation on physical manipulation
systems thus far. A large part of the robotic-reinforcement-learning (RL) community has
hence focused on simulation experiments 
\citep[e.g.][]{Haarnoja2018-ox, Fujimoto2018-ur, Popov2017-hh, Mnih2016-mn,
Heess2017-vd, duan2016benchmarking, henderson2018deep}. However, results
obtained in simulation experiments do often not translate to real systems
\citep[see e.g.][]{tobin2017domain, james2019sim}. The physics of contact
interaction are nonsmooth and the outcome is highly sensitive to parameters and
initial conditions (e.g. a slight difference in the shape of objects in
contact can lead to very different motions).

Therefore, we believe that an experimental platform, capable of generating large
amounts of data from a wide range of possible contact interactions with external
objects at a low cost, could greatly support progress in autonomous robotic
manipulation. 
The goal of this paper is to take a step in this direction. We present an
open-source robotic platform called TriFinger. Its hardware and software design
provide it with the following key strengths:
\begin{description}
	\setlength\itemsep{0.1em}
	\item[\namedlabel{req:sufficient_capabilities}{Dexterity}:] The robot design
	consists of three fingers and has the mechanical and sensorial capabilities
	necessary for complex object manipulation beyond grasping.
	\item[\namedlabel{req:unsupervised_execution}{Safe Unsupervised Operation}:] The
	combination of robust hardware and safety checks in the software allows users to
	run even unpredictable algorithms without supervision. This enables, for
	instance, training of deep neural networks directly on the real robot.
	\item[\namedlabel{req:ease_of_use}{Ease of Use}:]The C++ and Python interfaces
	are very simple and well-suited for reinforcement learning as well as optimal control
	at rates up to \SI{1}{kHz}. For convenience, we also provide a simulation
	(PyBullet) environment of the robot.
	\item[\namedlabel{req:viability}{Viability}:]The hardware design, based on
	\cite{Grimminger2020-tl}, is very simple and inexpensive (about
	\SI{5000}[\$]{} for the complete system), such that as many researchers as
	possible will be able to build their own platforms. All the information
	necessary for reproducing and controlling the platform is
	open-source\footnote{\label{google_site}\url{https://sites.google.com/view/trifinger}}.
\end{description}
An important point to note is that most of the software framework
is robot-agnostic and new systems can be integrated easily.
Hence, it is a contribution in its own right and can be used
independently of the robotic system proposed here.

In the remainder of the paper, we discuss related work and then describe the
design of the hardware and software in detail. Finally, we present experiments
in learning and optimal control to illustrate the aforementioned capabilities.

\section{Related Work}
In the past years, a large part of the RL community has focused on simulation
benchmarks, such as the deepmind control suite \cite{tassa2018deepmind} or
OpenAI gym \cite{brockman2016openai} and extensions thereof
\cite{zamora2016extending}. These benchmarks internally use physics simulators,
typically Mujoco \cite{todorov2012mujoco} or pyBullet
\cite{coumans2016pybullet}.

These commonly accepted benchmarks allowed researchers from different labs to
compare their methods, reproduce results, and hence build on each other's work.
Very impressive results have been obtained through this coordinated effort
\citep[see e.g.][]{Haarnoja2018-ox, Fujimoto2018-ur, Popov2017-hh, Mnih2016-mn,
Heess2017-vd, duan2016benchmarking, henderson2018deep}.

In contrast, no such coordinated effort was possible on real robotic systems,
since there is no shared benchmark. There have been isolated successes on real
systems \cite{levine2018learning, zhu2020the,
pinto2016supersizing,andrychowicz2020learning}, but these results cannot be
compared or reproduced, since each lab has their own robotic setup.

\subsection{Robot Hardware}
This lack of standardized real-world benchmarks has been recognized by the
robotics and reinforcement learning community \citep{behnke2006robot,
Bonsignorio2015-lg, Calli2015-vu, Calli2015-zj, Amigoni2015-gh, Murali2019-rg}.
Recently, there have been renewed efforts in this direction: For instance,
\citet{pickem2017robotarium} propose the \textsc{Robotarium}, a real-world
benchmark for mobile robots, and \citet{Grimminger2020-tl} propose an
open-source quadruped.
 
For manipulation, there are a number of affordable robots, such as Franka Emika,
Baxter, and Sawyer. \citet{yang2019replab} propose Replab, a simple manipulation
platform which has an even lower cost. Similarly, CMU proposed LoCoBot, a
low-cost, open-source platform for mobile manipulation. 
\citet{dieter} propose a design using pneumatic actuators and show
its suitability for robotic learning.
However, all of these
platforms have very simple end-effectors, typically 1-D grippers, which limit
the possibilities of interaction with the environment.

For dexterous manipulation, there are a number of robotic hands on the market
(e.g. BarrettHand, Shadow Hand, Schunk Hand) which cost typically at least
50\,000 Euro per hand (an affordable exception is the Allegro hand). 
In addition, \citet{dollar2010highly,she2015design,
xu2016design} proposed some innovative, exploratory hand designs. However, none
of these hands are designed for long-term unsupervised operation. In addition, to
have a sufficient workspace for manipulation, they have to be mounted on a robot
arm, which increases the complexity and risk of damage even further.

\begin{figure}
	\centering
	\includegraphics[height=0.26\textwidth]{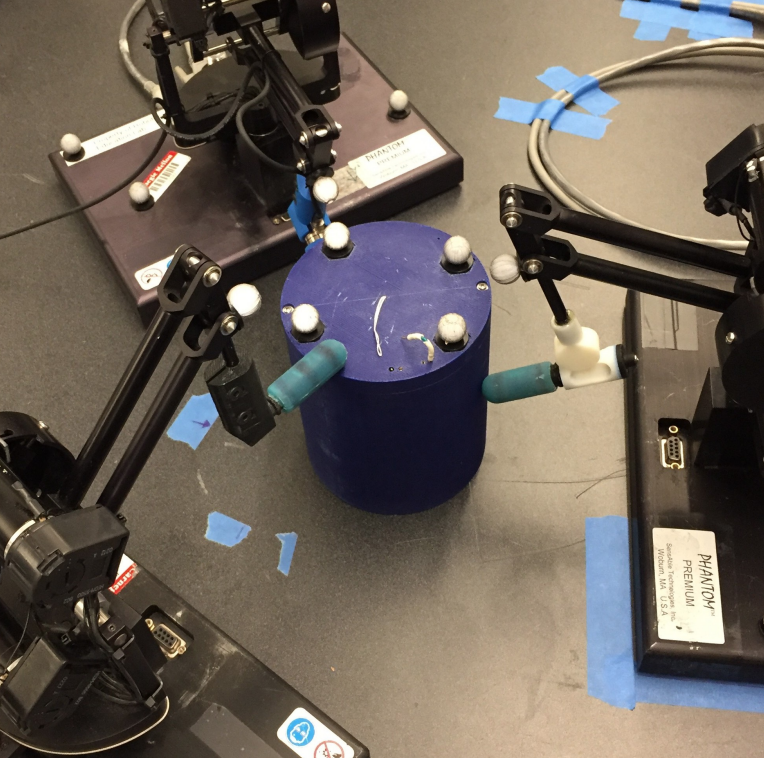}
	\includegraphics[height=0.26\textwidth]{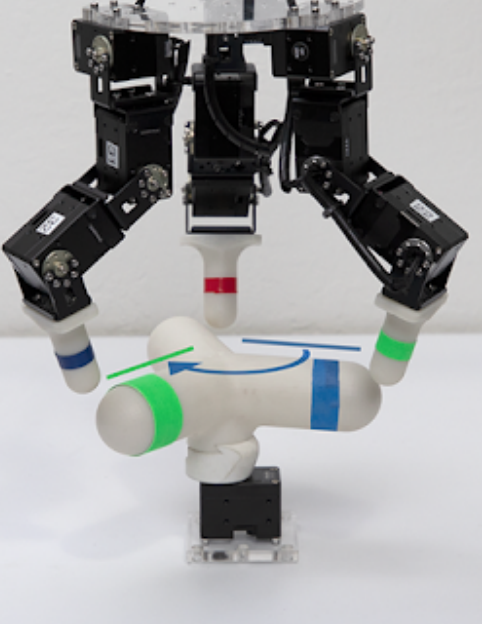}
	\caption{Phantom Manipulator (left, image from
		\cite{lowrey2018reinforcement})   and D'Claw (right, image from
		\cite{ahn2019robel}). As the TriFinger, both platforms consist of three
	manipulators with 3 DoF each.}
	\label{fig:competitors}
\end{figure}

The two setups which are most similar to the TriFinger are the D'Claw, a
three-fingered robotic hand \cite{ahn2019robel} and the Phantom Manipulation
Platform, consisting of three Phantom Haptic Devices
\cite{lowrey2018reinforcement}, see \cref{fig:competitors}. As the TriFinger,
both of these setups consist of three manipulators with 3 DoF each. However, the
workspace, where these manipulators can interact with objects, is much larger
for the TriFinger (see \cref{fig:workspace}). In addition, the Phantom
Manipulation Platform is at 30\,000\$ about six times more expensive than the
proposed setup. The D'Claw robot, at 3\,500\$, is in a similar price range as
the proposed setup, but its actuators allow for far less dynamic motion and are
less robust to impacts, as they are only backdrivable with substantial
force, i.e. they do not give in as easily. We provide more details on these points in 
\cref{appendix:comparison_d_claw}.
\subsection{Robot Software}
Robot software can roughly be divided into two classes:
\begin{itemize}
	\item Low-level software allows to communicate with the robot in real-time.
	      Essentially, it sends motor commands to the robot and retrieves sensory
	      measurements.
	\item High-level software which does not need to run in real-time and often
	      operates on a more abstract level. The most widely-used high-level software
	      is by far ROS. More recently  \citet{murali2019pyrobot} proposed PyRobot, a
	      python interface for motion generation and learning.
\end{itemize}%
Here we are mainly concerned with the low-level robot control software.
Unfortunately, there is no framework which is widely used across labs, instead
there is a large number of diverse solutions. These solutions typically rely on
programming languages close to machine language, in order to ensure that
control loops run in real-time (i.e. at a constant rate, e.g. \SI{1}{kHz}).
For example SL \cite{Schaal2009} uses C while ROS-control
\cite{ros_control}, LAAS-CNRS Stack-of-Tasks \cite{mansard:icar:09}, and
ETH control-toolbox \cite{adrlCT} are implemented in C++. A particular
case is the IHMC Robotics software \cite{ihmc} which uses a real-time Java with
a modified garbage collector. These frameworks differ in
how end-users integrate their controllers.  
SL and control-toolbox provide static methods or classes to
fill-in. ROS control uses the concept of ros-services to load and change
controllers online. Stack-of-Tasks relies on Python bindings to
dynamically interact with a control graph.

Unfortunately, these frameworks are only accessible to experienced users who
spent substantial amounts of time getting used to the particular approach at
hand. Implementation of controllers must usually follow a strict structure which
does not easily accommodate e.g. neural networks. In contrast, we designed the
user interface to be simple enough that even inexperienced users are able to
write controllers easily. It essentially consists of only two
functions: appending actions to a queue and accessing a history of observations.
Both functions are exposed in C++ as well as Python. The user may employ these
functions in any desired manner, for real-time or non-real-time control (using
e.g. a neural network). An additional benefit of the proposed design is that it
makes integration into high-level frameworks very simple.%
\section{Hardware Design}
\begin{figure}[ht]
	\begin{center}
		\includegraphics[width=0.8\columnwidth]{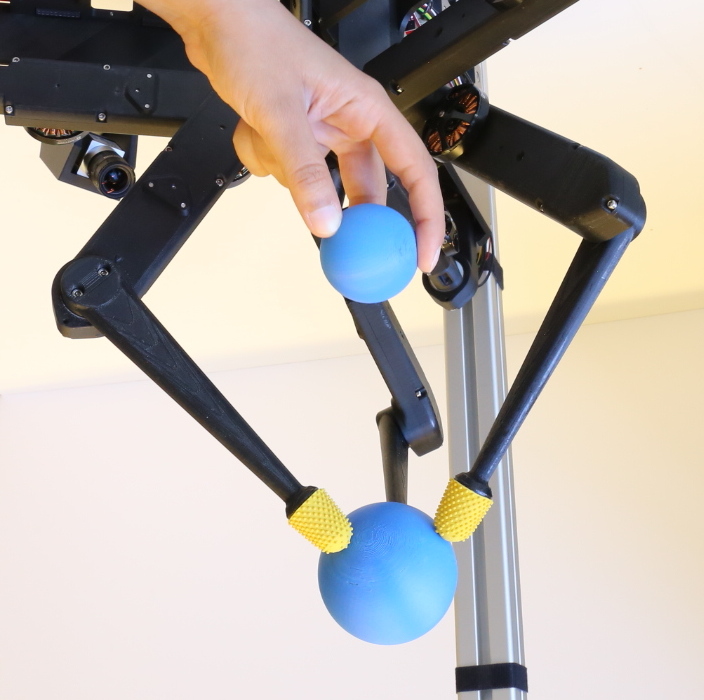}
		\caption{Analogy between the human hand and the proposed platform.}
		\label{fig:human_hand}
	\end{center}
\end{figure}

The hardware design is loosely inspired by thumb, index and middle finger of
a human hand, see \cref{fig:human_hand}.
We will therefore refer to the individual manipulators as fingers and to the
whole setup as TriFinger. In \cref{fig:catching} we can see  the main components
of the robotic platform, including the fingers, the frame and the three cameras.
In the following we will describe each of these components. All the details
necessary for building an instance of the proposed platform are open source
(\cref{google_site}).
\subsection{Finger Mechanics and Electronics}
\begin{figure}[ht]
	\begin{center}
		\includegraphics[width=0.75\columnwidth]{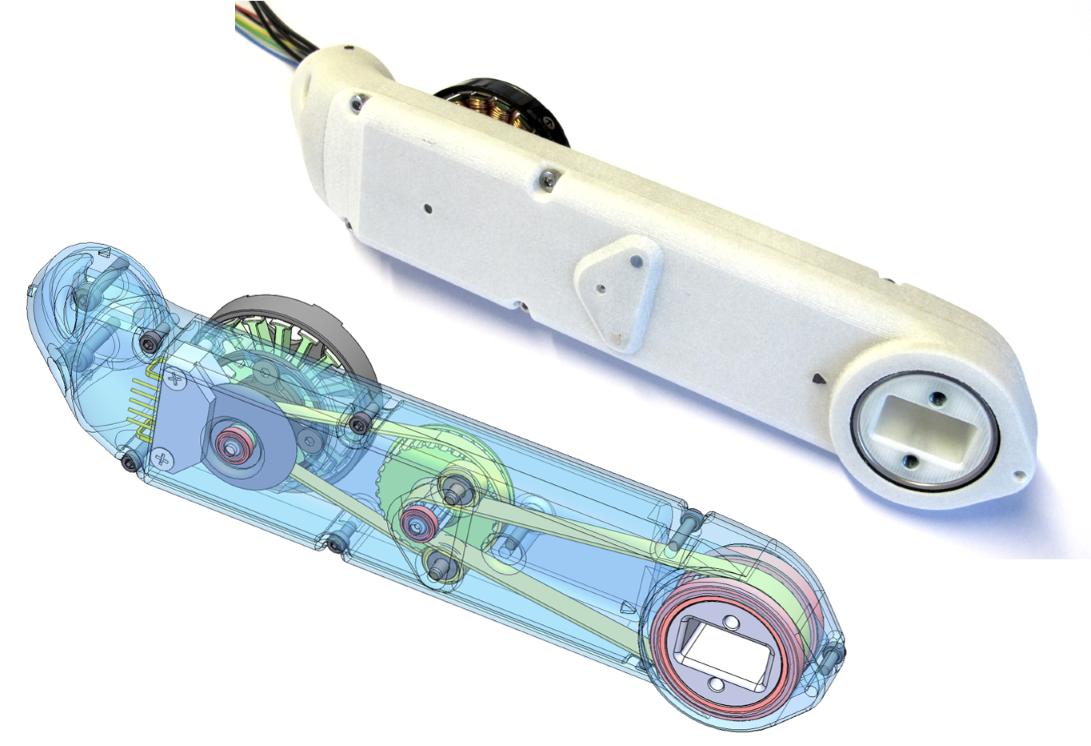}%
		\caption{Figure from \cite{Grimminger2020-tl} showing the actuator module which
		is the main building block of their quadruped legs and our fingers.}
		\label{fig:actuator_module}
	\end{center}
\end{figure}

The mechanics and electronics of the proposed robot are based on a recently
published open-source quadruped \cite{Grimminger2020-tl} consisting of
inexpensive high-performance motors, off-the-shelf parts, and 3D printed shells
(see \cref{fig:actuator_module}).
\begin{figure}[ht]
	\begin{center}
		\includegraphics[width=0.8\columnwidth]{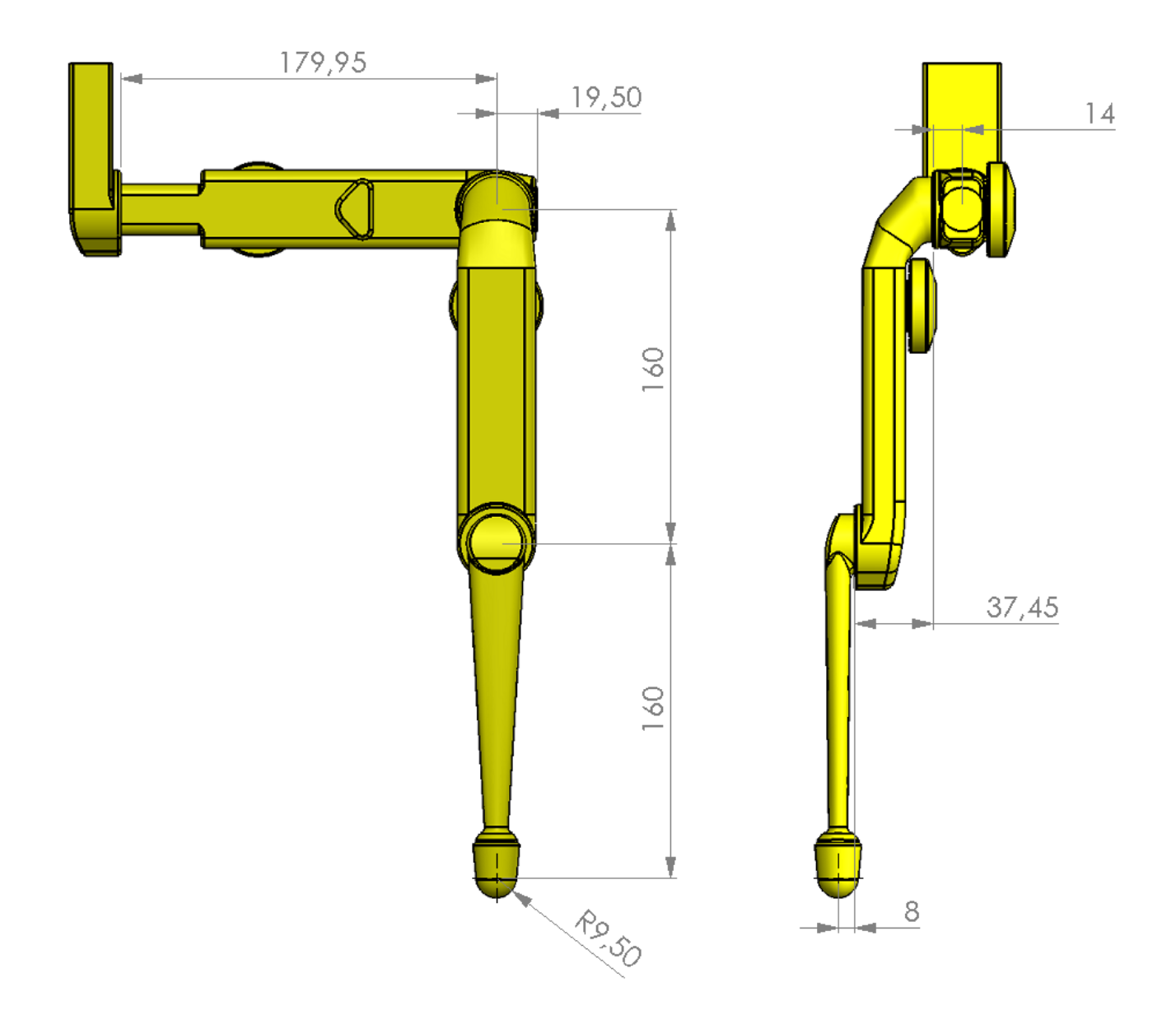}
		\caption{Technical drawing of a single finger, measurements are in mm.}
		\label{fig:technical_drawing}
	\end{center}
\end{figure}%
\citet{Grimminger2020-tl} originally proposed a 2 DoF leg, which they extended to
3 DoF in the meanwhile (see
\footnote{\url{https://github.com/open-dynamic-robot-initiative/open_robot_actuator_hardware/tree/master/mechanics}}
for an overview of the designs based on \cite{Grimminger2020-tl}). The fingers
of the proposed platform are identical to the 3 DoF version of the quadruped
leg, apart from some slight modifications to the mounting, the 3D-printed shells, and
the end-effector.
We hence inherit all the favorable properties of the design from
\citet{Grimminger2020-tl}:
\begin{itemize}
	\item The high-performance brushless DC motors provide \textbf{high-torque}
	      actuation while having a \textbf{low weight}, which allows for dynamic
	      manipulation of objects up to a few hundred grams in weight.
	\item Transparency of the transmission enables \textbf{force control and
		      sensing} and \textbf{robustness to impacts}, both of which are crucial for
		robotic manipulation. Transparency means that forces applied at the
		end-effector directly translate to torques at the motors, rather than being
		absorbed by a high-gear-ratio transmission. This implies that end-effector
		forces can be obtained by measuring the motor currents and that impacts will
		not break the transmission.
		\item The motors can be \textbf{controlled at high frequency (1 kHz)} from a
		      consumer computer with a realtime-patched Ubuntu. This allows the robot to
		      sense external forces and react to them extremely quickly.
		\item The \textbf{design is very simple} and conists of \textbf{inexpensive}
		      off-the-shelf parts and 3D printed shells. This will allow other researchers
		      to build their own platforms.
	\end{itemize} 
	A small but important addition in our design is a soft tip which mimics the
	human finger tip. This increases the stability of interaction with external
	objects greatly, as impacts are damped and contact extends to a surface rather
	than a single point.
	\subsection{Kinematics}
	\begin{figure}[ht]
		\begin{center}
			\includegraphics[width=0.8\columnwidth]{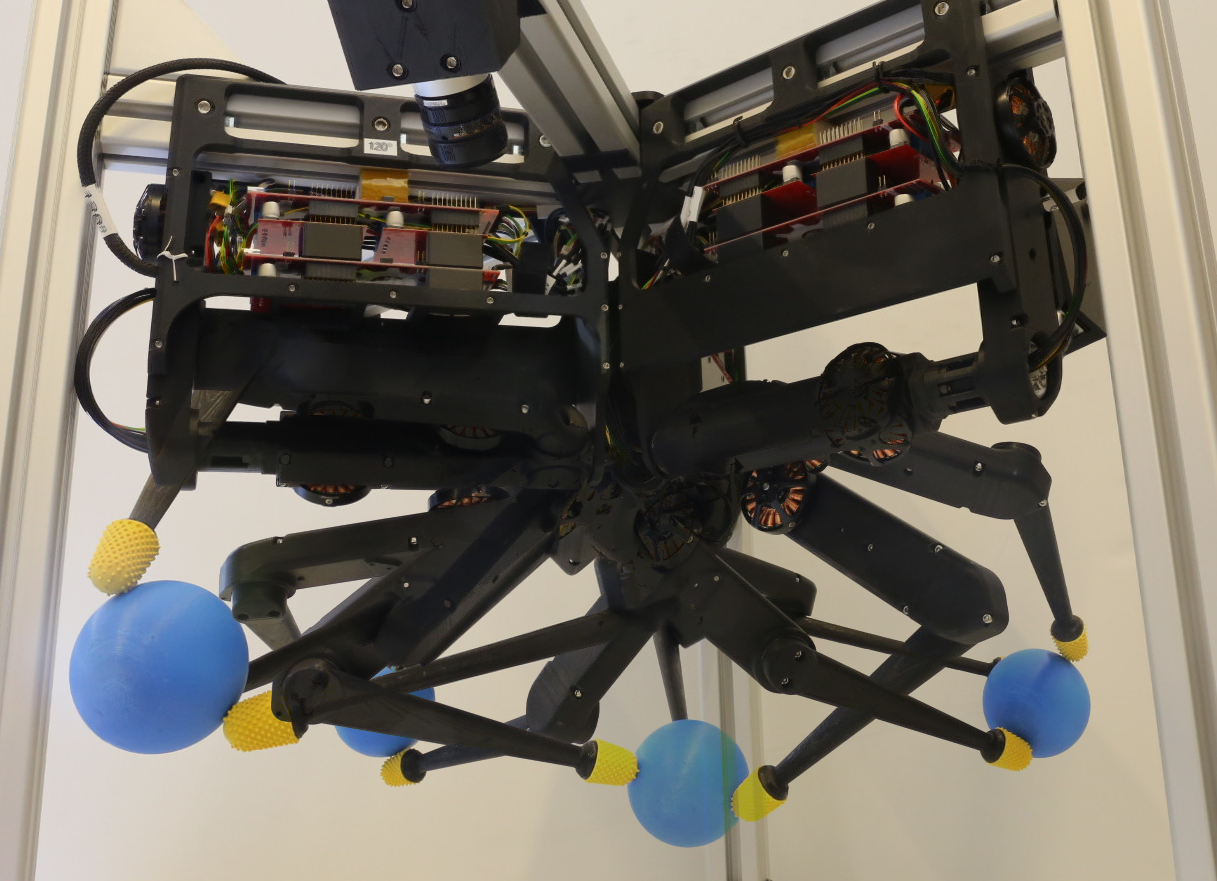}
			\caption{An overlay of different robot configurations to illustrate the workspace.}
			\label{fig:workspace}
		\end{center}
	\end{figure}%
	The kinematics were designed to maximize the workspace where all three fingers
	can interact with an object simultaneously, see \cref{fig:workspace}.

	Each of the manipulators has 3 DoF, which implies that the finger tip can move
	in any direction (see \cref{fig:technical_drawing} for a technical drawing). This is
	important for dexterity and it makes the finger robust to impacts, as it can
	give in to forces from any direction.
	\subsection{Frame and Boundary}
	\begin{figure}[ht]
		\begin{center}
			\includegraphics[width=0.8\columnwidth]{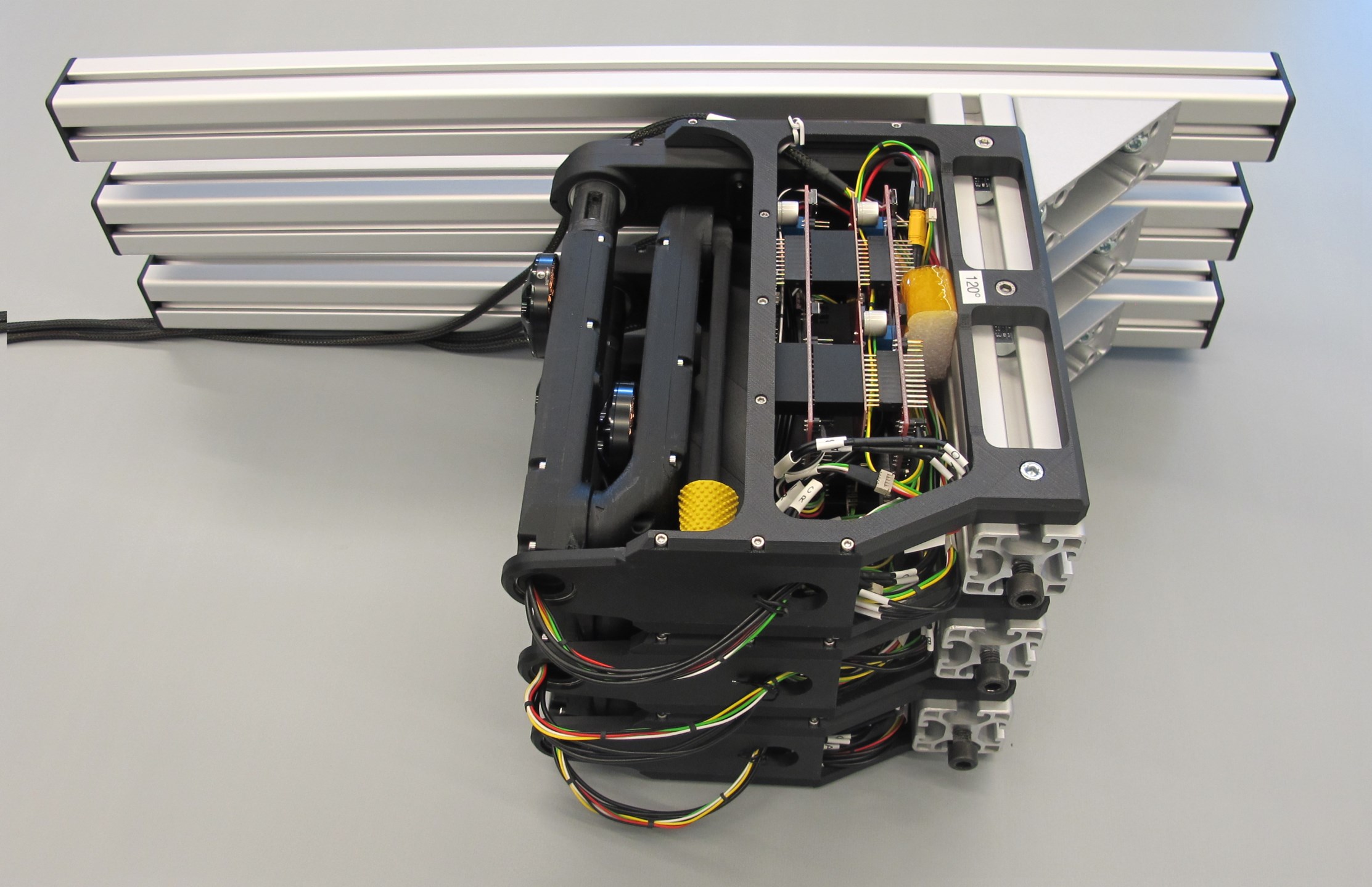}
			\caption{The platform can be disassembled into three modules for  transportation and storage.}
			\label{fig:disassembled}
		\end{center}
	\end{figure}%
	We use an aluminium frame to attach the fingers and cameras. The height of the
	fingers can be adjusted easily according to the requirements of a specific task.
	In addition, the entire platform can simply be flipped (see \cref{fig:catching})
	for tasks such as throwing and catching.
			
\begin{figure}[ht]
		\begin{center}
			\centerline{\includegraphics[width=0.8\columnwidth]{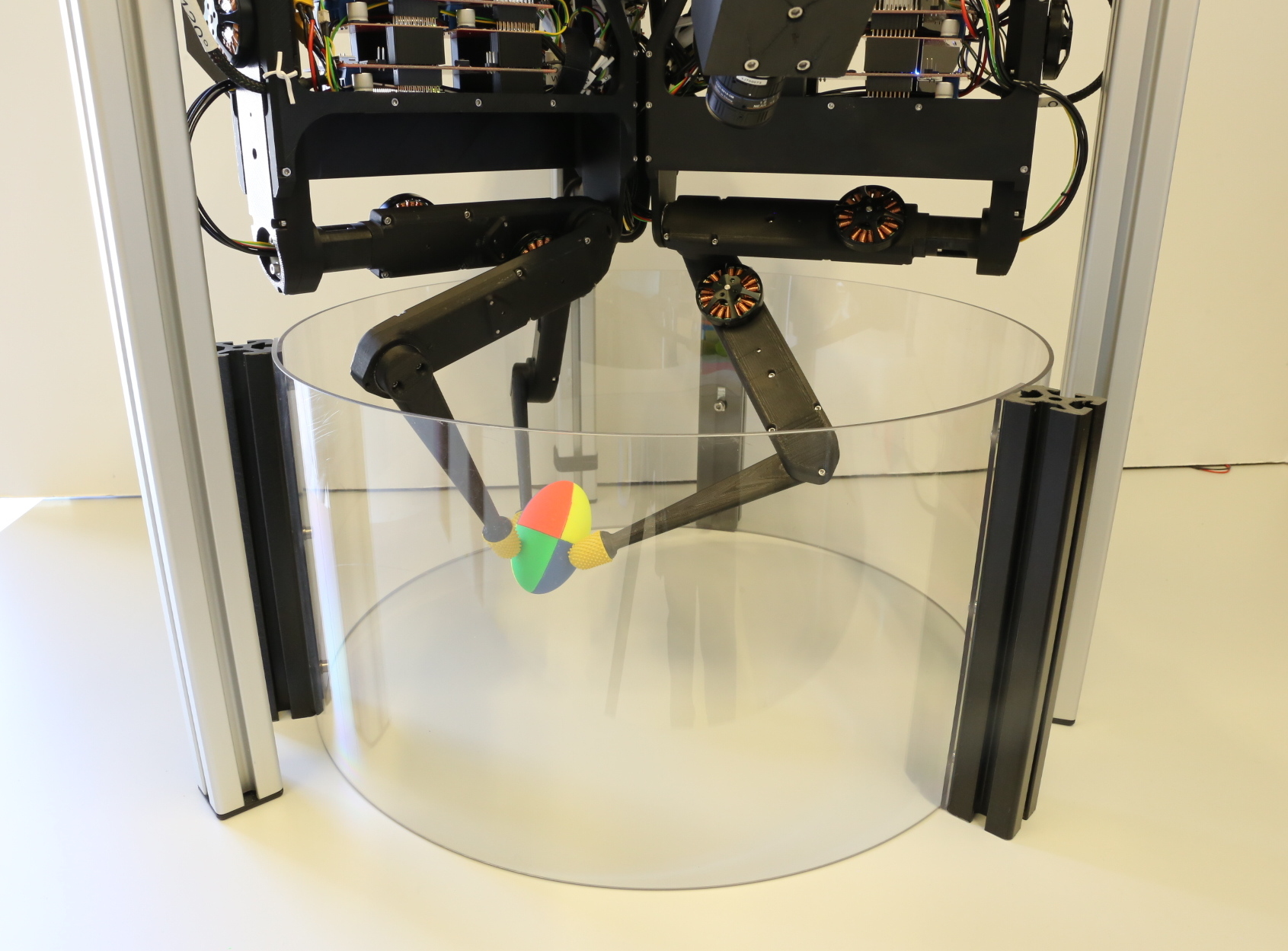}}
			\caption{A boundary to prevent objects from leaving the workspace.}
			\label{fig:high_arena_border}
		\end{center}
	\end{figure}

	For manipulation on the table, we designed an optional boundary to confine
	objects to the workspace of the platform, see \cref{fig:high_arena_border}. This
	is essential for learning during extended periods of time without human
	supervision.
			
	Finally, the platform can be disassembled easily into three compact modules, see
	\cref{fig:disassembled}.
	\subsection{Cameras}
\begin{figure}[ht]
		\begin{center}
			\centerline{
				\includegraphics[width=0.32\columnwidth]{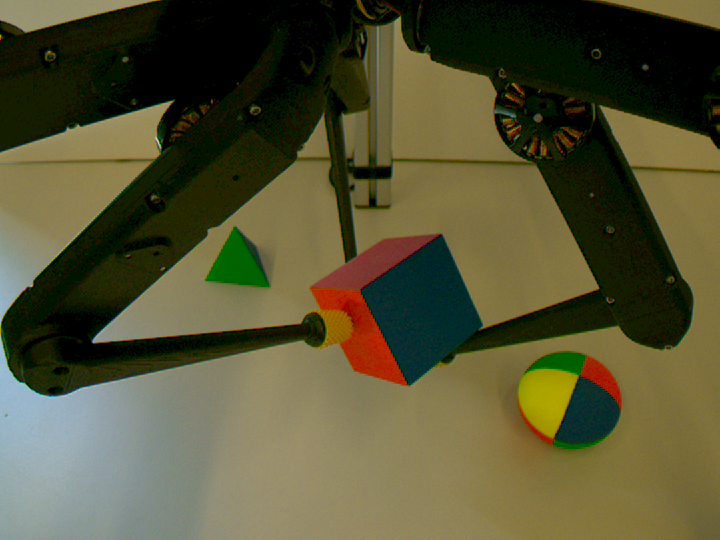}
				\includegraphics[width=0.32\columnwidth]{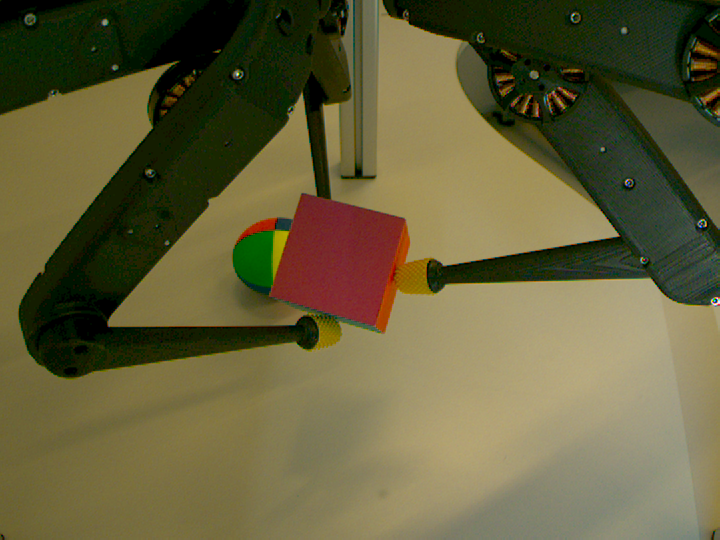}
				\includegraphics[width=0.32\columnwidth]{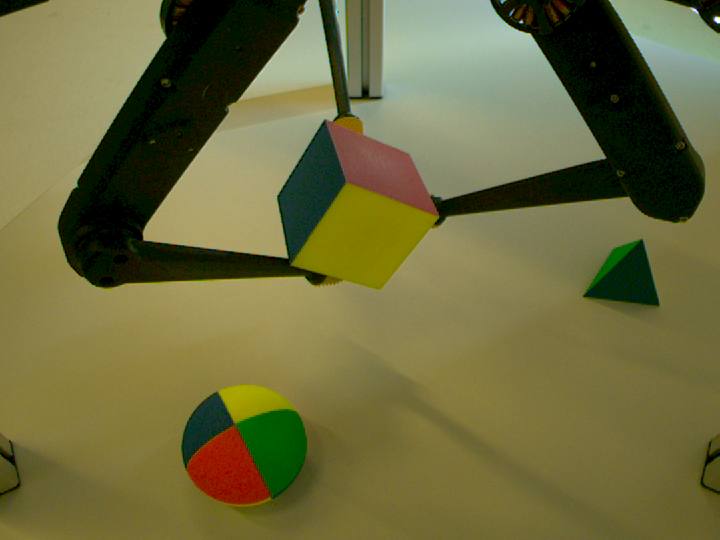}
			}
			\caption{Images taken from each of the three cameras. Having three cameras ensures that the object will always be visible from at least one camera.}
			\label{fig:cameras}
		\end{center}
	\end{figure}

	As there are three fingers interacting closely with the target object, there
	is a lot of potential for occlusion. Therefore we place three cameras around
	the platform (see \cref{fig:catching}), ensuring that the object will at all
	times be visible from at least one camera. We use Basler acA720-520uc
	cameras with Basler C125-0418-5M-2000034830 lenses, as they have a high rate
	of up to 525 fps using global shutter, low latency, and an appropriate field
	of view (see \cref{fig:cameras}).
	\subsection{Measurements and Control Signals}
	As in \cite{Grimminger2020-tl}, the signals are transferred between the control
	computer (standard consumer PC) and the motors through CAN at 1 kHz. In
	accordance with RL terminology, we will call the control signal action and the
	measurements observation.\\
	\textbf{Action:} The input of this platform is a
	nine-dimensional vector: A desired torque (which is proportional to the
	current) for each joint. The robot expects this signal to be sent at a rate
	of 1 kHz.\\
	\textbf{Observation:} The output consists of proprioceptive measurements
	(joint angles, joint velocities, joint torques), acquired at 1 kHz, and
	images from the three cameras, obtained typically at 100 Hz.

	\section{Software Design}
	The key strengths of our software framework are:
	\begin{itemize}
		\item The user interface in C++ and Python is very \textbf{simple}, yet
		      well-suited for real-time (1 kHz) optimal control and reinforcement
		      learning.
		\item It performs \textbf{safety checks} to prevent the robot from breaking.
		      This frees the user from this burden and allows them to execute even complex and
		      unpredictable algorithms without surveilling the platform. This opens, for
		      instance, the possibility of training a deep neural network policy during
		      several days directly on the robot.
		\item A synchronized \textbf{history of all the inputs and outputs} of the
		      robot is available to the user and can be logged.
		\item The software is designed such that new robots and simulators can
		      easily be integrated. This may facilitate reuse of algorithms across robots.
	\end{itemize}
	\subsection{Control Modes}
	There are two modes of control supported by our software:
	\begin{definition}[Real-time control]\label{def:real_time_control} By real-time
		control we mean that actions have to be sent to the system at a fixed rate of
		$\Delta$ seconds. This is necessary for real-world dynamical systems, as they
		evolve in time and can hence not wait for the next control to arrive. For
		instance, a falling humanoid robot cannot stand still in the air to wait for
		computation of the next action to be completed.
	\end{definition}%
	\begin{definition}[Non-real-time control]\label{def:non_real_time_control} By
		non-real-time control we mean that a change in the time at which actions are
		applied does not change the outcome, and that actions may take varying amounts
		of time. An example of such a system is a simulator: it will wait until the next
		action is provided and simulation of an action may take varying amounts of time
		for computational reasons. Another example is a a mobile robot which is
		controlled through highlevel actions, such as moving to a specified goal
		location. In between actions the robot will stand still, and execution time of
		an action will vary according to the distance to the goal etc.
	\end{definition}%
	An important feature of our software design is that both modes are supported
	through the same interface, which makes it easy to run the same code in
	simulation and on the real robot.
	\subsection{Overview}
	\begin{figure}[ht]
		\begin{center}
			\centerline{\includegraphics[width=0.9\columnwidth]{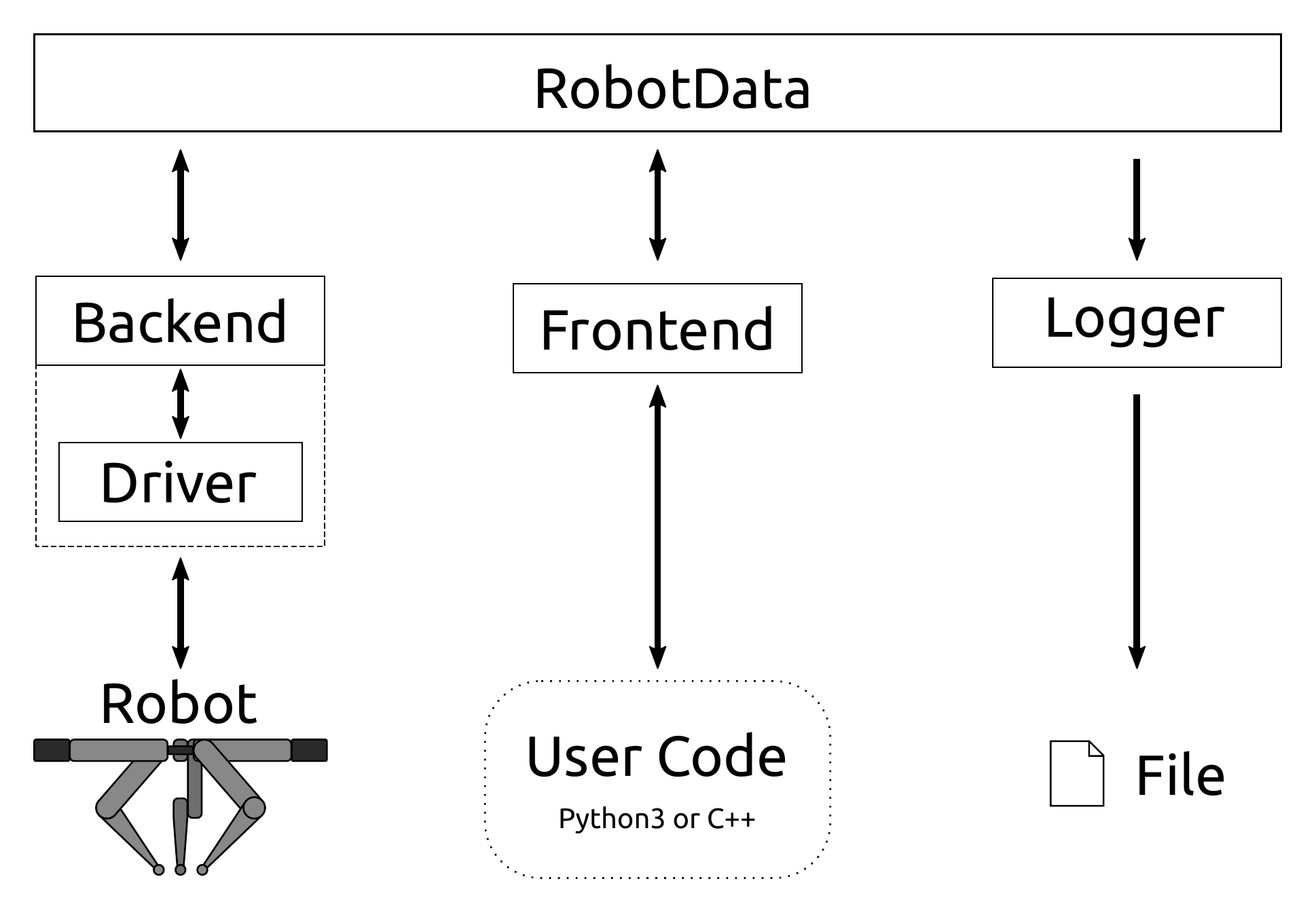}}
			\caption{Software Architecture.  The modules communicate through the RobotData and do not have any direct connections.}
			\label{fig:software_architecture}
		\end{center}
	\end{figure}
	The software framework has three main components (see
	\cref{fig:software_architecture}):
	\begin{itemize}
		\setlength\itemsep{0.1em}
		\item The \textbf{back-end} communicates with the robot through the driver,
		\item the \textbf{front-end} allows the user to control the robot through
		      C++ or Python3,
		\item the \textbf{logger} logs all the inputs and outputs of the robot.
	\end{itemize}
	Each of these components can run in a separate process, which is advantageous
	for computational reasons and it separates the back-end from the user code.
	Since all the commands sent to the robot flow through the back-end, we can
	implement checks which ensure safety no matter what happens in the user code.
			
	In the following we will describe the front-end and the back-end. The back-end
	section is relevant for readers interested in setting up their own robot, users
	only need to know about the front-end.
	\subsection{Front-end}
\begin{figure}[ht]
		\begin{center}
			\centerline{\includegraphics[width=0.9\columnwidth]{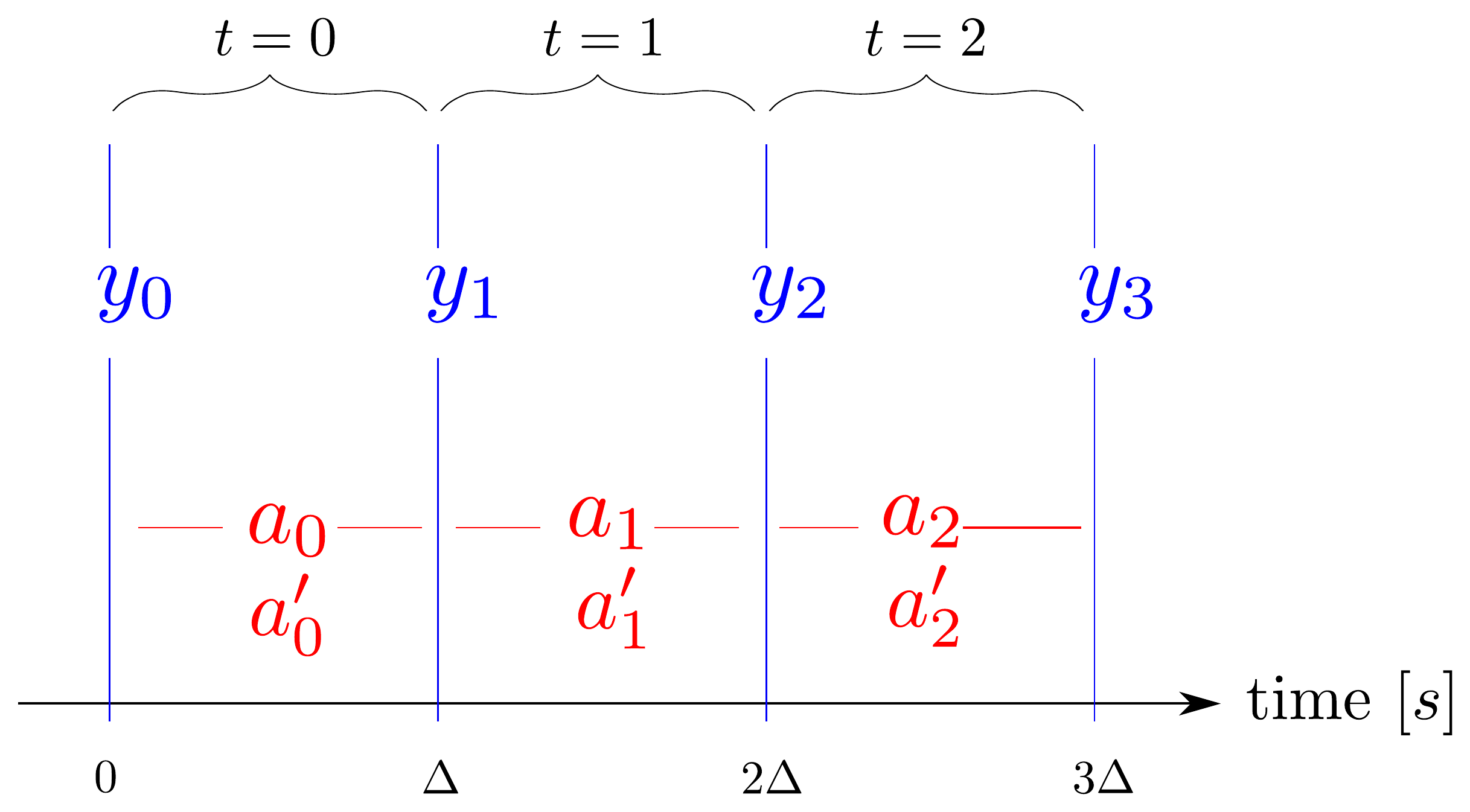}}
			\caption{This figure shows the temporal sequence of variables. In real-time mode
				(\ref{def:real_time_control}) there is a fixed control and observation rate
				$\Delta$. Each observation $y_t$ corresponds to a single instant in time
				$t\Delta$, while each action $a_t$ corresponds to a time interval $[t \Delta ,
					(t+1)\Delta)$.}
				\label{fig:variables_sequence}
			\end{center}
		\end{figure}

	In our design, the only objects that exist from the user-perspective are
	\begin{itemize}
		\setlength\itemsep{0.1em}
		\item a time-series of desired actions $a$ computed by the user,
		\item a time-series of actions actually applied to the robot $a'$, which is
		      identical to $a$ except for potential modifications to satisfy safety
		      constraints,
		\item and a time-series of observations $y$.
	\end{itemize}
	\Cref{fig:variables_sequence} shows the temporal relations of these variables.
		The user may perform two operations: They may append actions to the
		desired-action time-series $a$ and they may read from any of the three
		time-series $a, a', y$, where $a',y$ are filled-in by the back-end.
						
		The user interface implementation is equivalent to the following pseudo code:
		{\small
			\setlength{\abovedisplayskip}{6pt}
			\setlength{\belowdisplayskip}{\abovedisplayskip}
			\setlength{\abovedisplayshortskip}{0pt}
			\setlength{\belowdisplayshortskip}{3pt}
			\begin{align*}
				\texttt{def } & \texttt{append\_desired\_action(\ensuremath{x})}:               \\
				              & \texttt{\ensuremath{a\leftarrow(a,x)}}                          \\
				              & \texttt{\texttt{return len(\ensuremath{a})}-1}                  \\
				\texttt{def } & \texttt{get\_observation(\ensuremath{t})}:                      \\
				              & \texttt{wait until len(\ensuremath{y})>\ensuremath{t}}          \\
				              & \texttt{return }y_{t}                                           \\
				\texttt{def } & \texttt{get\_desired\_action(\ensuremath{t})}:\texttt{as above} \\
				\texttt{def } & \texttt{get\_applied\_action(\ensuremath{t})}:\texttt{as above} 
			\end{align*}
		}
		This interface provides access to a synchronized history of all the inputs and
		outputs of the robot. The user has complete freedom how to use this data and
		when to append actions, as long as they make sure to do so on time, before the
		action is needed by the robot. For instance, they may choose to run a typical
		real-time control loop where a new action $a_t$ is computed periodically, or
		they may choose to compute entire action sequences at a lower rate and then
		append them in a burst by repeatedly calling \texttt{append\_desired\_action}.
						
		If the user attempts to access a future observation through
		\texttt{get\_observation($t$)}, this function will wait and return as soon as
		this observations is acquired. For instance, in the case of real-time control
		(\cref{def:real_time_control}), if the call \texttt{get\_observation($2$)} is
		made at time $<2\Delta$ seconds, the function will wait until the observation
		$y_2$ is received at time $2\Delta$ seconds and then return. This feature allows
		for synchronization with the real system. 
						
		The function \texttt{append\_desired\_action($x$)} will append action $x$ to the
		action time-series. For convenience it returns the timeindex of the appended
		action, but if the user keeps track of timeindices externally, the return value
		can be ignored. The instant of the first call to
		\texttt{append\_desired\_action} marks time $0$, after which the back-end will
		start filling in the $a', y$ timeseries and expect the action timeseries $a$ to
		be filled in by the user.
						
		A basic control loop can be written as follows:
		\begin{example}[Basic control loop]\label{ex:basic_control_loop} {\small
				\setlength{\abovedisplayskip}{6pt}
				\setlength{\belowdisplayskip}{\abovedisplayskip}
				\setlength{\abovedisplayshortskip}{0pt}
				\setlength{\belowdisplayshortskip}{3pt}
				\begin{align*}
					  & \texttt{robot.append\_desired\_action(\ensuremath{a_{0}})}                 \\
					  & \texttt{for \ensuremath{t} in \ensuremath{(0,...,T)}:}                     \\
					  & \qquad\texttt{\ensuremath{y_{t}} = robot.get\_observation(\ensuremath{t})} \\
					  & \qquad\texttt{\ensuremath{a_{t+1}} = some\_policy(\ensuremath{y_{t}})}     \\
					  & \qquad\texttt{robot.append\_desired\_action(\ensuremath{a_{t+1}})}         
				\end{align*}}
			No explicit wait is necessary, synchronization with the back-end is ensured
			through the call to \texttt{robot.get\_observation(\ensuremath{t})}, which will
			wait until the back-end has appended $y_t$.
									
			This control loop is valid for both real-time (\ref{def:real_time_control}) and
			non-real-time (\ref{def:non_real_time_control}) control:\\
			If the back-end is running in \textbf{real-time mode}, it will add observations
			$y_t$ periodically and the loop above will hence run at a fixed rate, unless the
			call \texttt{some\_policy} is too slow. If that is the case, the back-end will
			detect that the user did not append the next action on time and it will shut
			down the robot and raise an error. Alternatively, the back-end can be configured
			to simply repeat the previous action if the next action has not been appended on
			time.\\
			In contrast, if the back-end is running in \textbf{non-real-time
			mode}, it will wait for the next action, and the function
			\texttt{append\_desired\_action} may be called with arbitrary delay.
			Similarly, the execution of actions may take varying amounts of
			time, and hence the call to \texttt{get\_observation} will not
			return at a predetermined time. This mode makes sense e.g. for
			simulation, where the simulator and the policy add actions and
			observations whenever they are done with their respective
			computations.
		\end{example}
						
		If we wish to control the robot at a lower rate, this can also be implemented
		very easily:
		\begin{example}[Basic control loop at a reduced
			rate]\label{ex:basic_control_loop_at_lower_rate} {\small
				\setlength{\abovedisplayskip}{6pt}
				\setlength{\belowdisplayskip}{\abovedisplayskip}
				\setlength{\abovedisplayshortskip}{0pt}
				\setlength{\belowdisplayshortskip}{3pt}
				\begin{align*} & \texttt{robot.append\_desired\_action(\ensuremath{a_{0}})}\\
					  & \texttt{for \ensuremath{t} in \ensuremath{(0,...,T)}:}                                         \\
					  & \qquad\texttt{\ensuremath{y_{t}} = robot.get\_observation(\ensuremath{t}\ensuremath{\cdot k})} \\
					  & \qquad\texttt{\ensuremath{a_{t+1}} = some\_policy(\ensuremath{y_{t}})}                         \\
					  & \qquad\texttt{for \_ in range(\ensuremath{k})}                                                 \\
					  & \qquad\texttt{\ensuremath{\qquad}robot.append\_desired\_action(\ensuremath{a_{t+1}})}          
				\end{align*}}
			Here, the control frequency is reduced by a factor of $k$. Note that the
			index $t$ refers to the control cycle here, not the robot cycle.
		\end{example}
		\subsection{Relation to the Standard Reinforcement Learning Framework}
		As we shall see, there is a gap between the standard Markov decision process
		(MDP) formulation of reinforcement learning and the presented framework. This is
		due to the realtime constraints we face when working with real robots. Here we
		show show to close this gap, which allows us to e.g. wrap our interface into
		the OpeanAI Gym interface.
						
		\paragraph{Some Simplifications:} For simplicity of exposition, we will only
		consider the case where the observations $y_t$ are Markovian, the extension to
		the non-Markovian case is straightforward. In addition, we will not treat the
		applied actions $a'_t$ explicitly anymore, as they do not affect the independence
		structure between the desired actions $a_t$ (henceforth referred to simply as
		actions) and observations $y_t$. They can either be simply ignored or they can
		be added to the observations if one wishes to give the RL algorithm access to
		this information. 
						
		\subsubsection{Dependence Structure of a Real-Time System} 
\begin{figure}[tb]
			\begin{center}
				\centerline{\includegraphics[width=0.5\columnwidth]{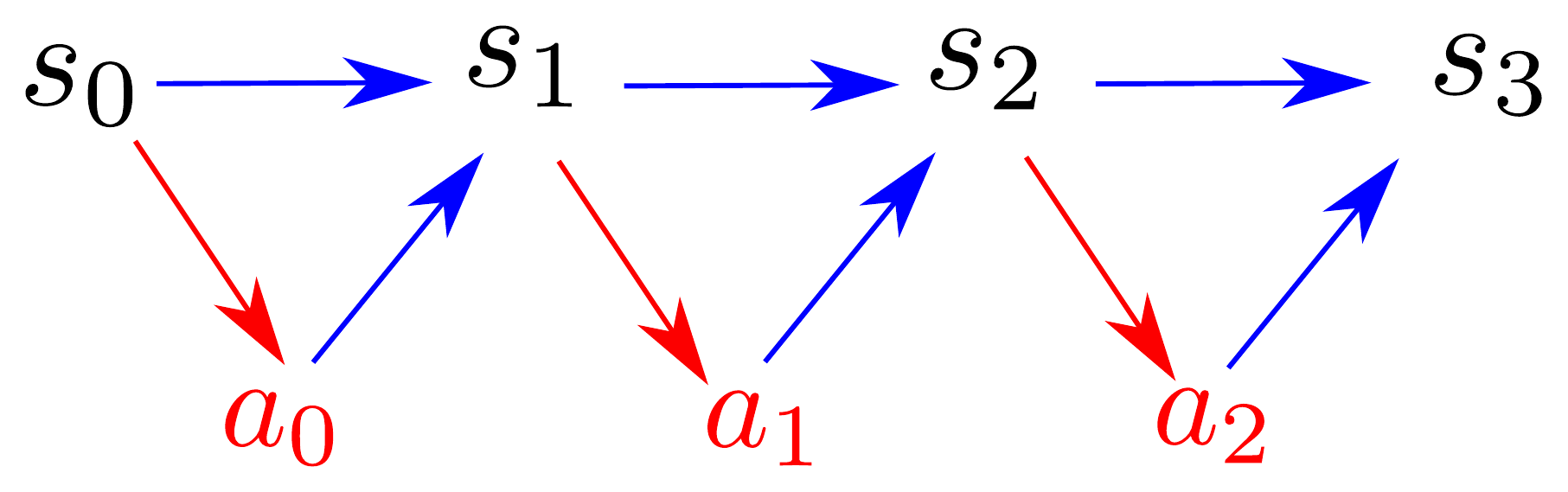}}
				\caption{The dependence structure (Bayes net) of an MDP.}
				\label{fig:variables_dependencies_gym}
			\end{center}
		\end{figure}

		We can deduce the dependences between variables from their temporal ordering
		represented in \cref{fig:variables_sequence}. Naturally, any variable can only
		causally depend on variables which precede it in time. However, in real-time
		systems there is the additional constraint that the controller needs some time
		to compute actions, hence they may not depend on observations they coincide with
		in time. More precisely, action $a_t$ cannot depend on $y_t$, since $a_t$ starts
		at precisely the time when observation $y_t$ is acquired. In fact, we already
		took this into account in \cref{ex:basic_control_loop}, where $a_{t+1}$ is a
		function of $y_t$ rather than $y_{t+1}$.
		\begin{figure}[tb]
			\begin{center}
				\centerline{\includegraphics[width=0.5\columnwidth]{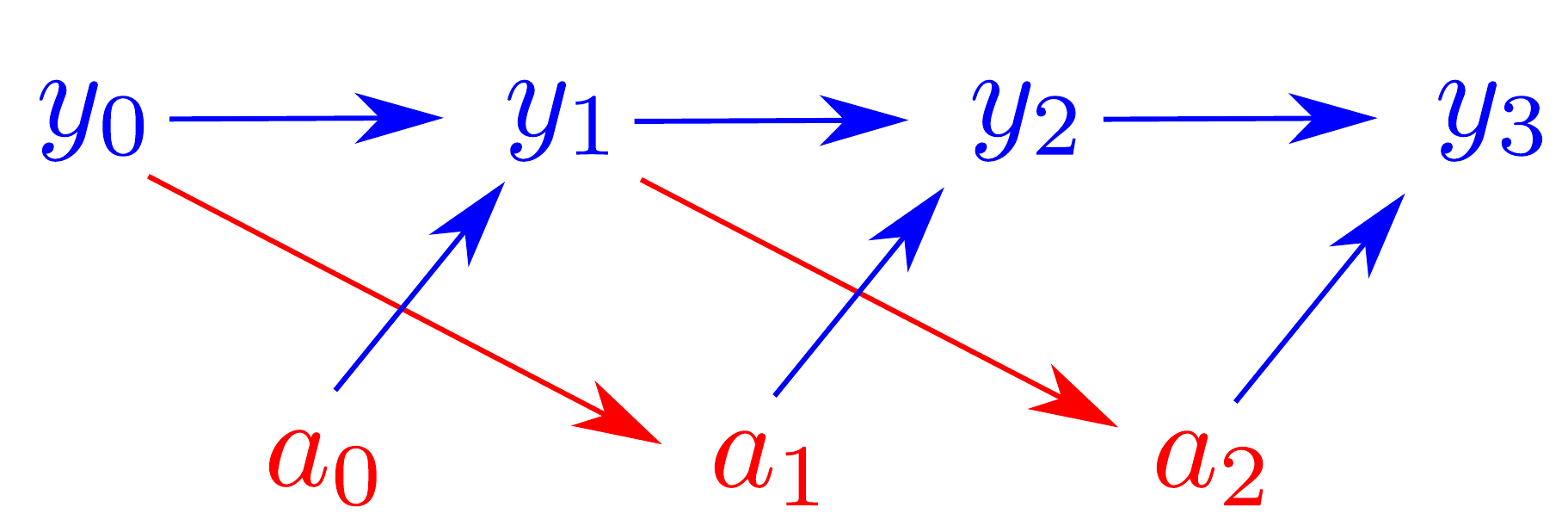}}
				\vspace{-0.1cm}\caption{\vspace{-0.1cm}The Bayes net showing the dependence
				between variables in a real-time system.}
				\label{fig:variables_dependencies_real}
			\end{center}
		\end{figure}
		These considerations lead to the Bayes net in
		\cref{fig:variables_dependencies_real}. By comparing with the Bayes net of an
		MDP in \cref{fig:variables_dependencies_gym} we see that we cannot simply
		identify the state $s_t$ with $y_t$, even in the Markovian case. In the
		following, we propose two ways in which this structure can be mapped to a
		standard MDP.
		\subsubsection{State Augmentation}
\begin{figure}[tb]
			\begin{center}
				\centerline{\includegraphics[width=0.5\columnwidth]{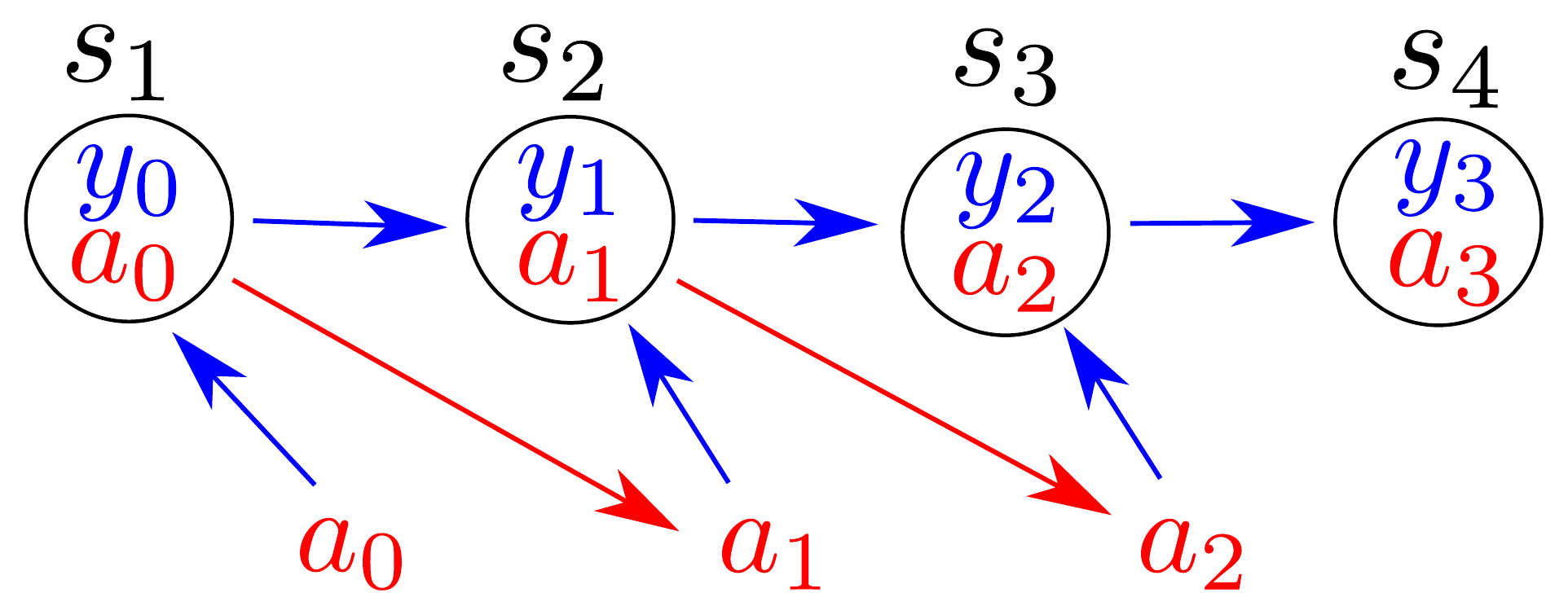}}
				\caption{A Bayes net showing the dependence between variables obtained from
					\cref{fig:variables_dependencies_real} by defining the state as
					$s_{t+1}=(y_t , a_t)$. This yields the same dependency structure as in
					\cref{fig:variables_dependencies_gym}.}
				\label{fig:variables_dependencies_compromise}
			\end{center}
		\end{figure}

		By defining the state in the slightly counter-intuitive way $s_{t+1} = (y_{t},
		a_{t})$, we retrieve the dependence structure of the MDP, see
		\cref{fig:variables_dependencies_compromise}. 
		This allows us to apply any of the RL algorithms which have been formulated for
		standard MDPs to our system. \citet{ramstedt2019real} use a similar trick and
		propose a method to take the resulting structure into account when performing
		RL.
						
		Given these considerations, we can now define an OpenAI Gym environment for our
		robot interface. For our discussion here, only the step function is relevant:
		\begin{example}[Gym Env - State
			Augmentation]\label{para:gym_state_augmentation}
			{\small
				\setlength{\abovedisplayskip}{6pt}
				\setlength{\belowdisplayskip}{\abovedisplayskip}
				\setlength{\abovedisplayshortskip}{0pt}
				\setlength{\belowdisplayshortskip}{3pt}
				\begin{align*}\texttt{def } & \texttt{step(\ensuremath{a}):}\\
					  & \texttt{\ensuremath{t} = robot.append\_desired\_action(\ensuremath{a})} \\
					  & \texttt{\ensuremath{y} = robot.get\_observation(\ensuremath{t})}        \\
					  & \texttt{\ensuremath{s'} = \ensuremath{(y,a)}}                           \\
					  & \texttt{return \ensuremath{s'}, some\_other\_stuff}                     
				\end{align*}
			} Where the $'$ in $s'$ denotes the subsequent time index.
		\end{example}
		This opens up the possibility of applying numerous implementations of RL
		algorithms out-of-the-box to our platform. For instance, we can directly apply
		the algorithms from \cite{stable-baselines}, \cite{garage} and \cite{baselines}, as we will demonstrate in the experimental section.
						
		It is often the case that we want to operate at a control rate which is lower
		than the communication rate of the robot. For instance, in many tasks it is not
		necessary to control the proposed system at $1000$ Hz, we may want to control it
		at e.g. $100$ Hz. We can define an according OpenAI Gym environment.
		\begin{example}[Gym Env - State Augmentation at Reduced
			Rate]\label{para:gym_state_augmentation_reduced} The logic here is exactly the
			same as in \cref{para:gym_state_augmentation}, except that each action is
			applied $k$ times, where $k$ is the rate reduction factor. {\small
				\setlength{\abovedisplayskip}{6pt}
				\setlength{\belowdisplayskip}{\abovedisplayskip}
				\setlength{\abovedisplayshortskip}{0pt}
				\setlength{\belowdisplayshortskip}{3pt}
				\begin{align*}\texttt{def } & \texttt{step(\ensuremath{a}):}\\
					  & \texttt{\ensuremath{t} = robot.append\_desired\_action(\ensuremath{a})}   \\
					  & \texttt{for \_ in range(\ensuremath{k-1\text{):}}}                        \\
					  & \texttt{\ensuremath{\qquad}robot.append\_desired\_action(\ensuremath{a})} \\
					  & \texttt{\ensuremath{y} = robot.get\_observation(\ensuremath{t})}          \\
					  & \ensuremath{s'}\texttt{ = }\ensuremath{(\ensuremath{y},\ensuremath{a})}   \\
					  & \texttt{return \ensuremath{s'}, some\_other\_stuff}                       
				\end{align*}}
		\end{example}
		\subsubsection{Approximate Mapping for Control at Lower Rates}
\begin{figure}[tb]
			\begin{center}
				\centerline{\includegraphics[width=0.95\columnwidth]{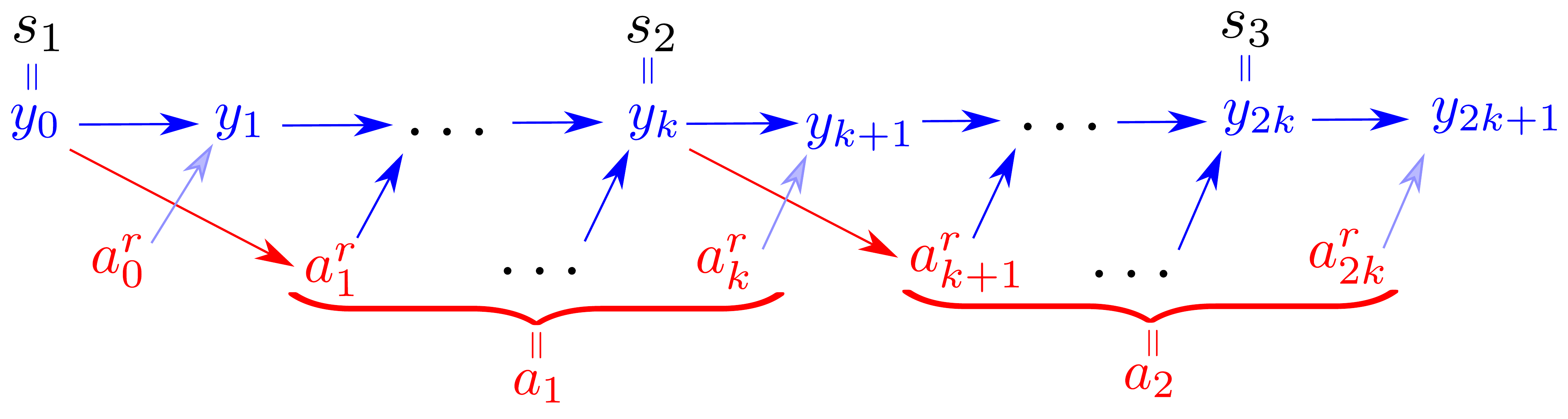}}
				\caption{When we control at a reduced rate each control action
				$a_t$ is applied $k$ times and hence produces $k$ robot actions
				$a^r_{(t-1)k+1:tk}$. To approximately map this to the dependence
				structure of a standard MDP
				(\cref{fig:variables_dependencies_gym}), we can ignore the
				light-blue arrows.}
				\label{fig:variables_dependencies_approximation}
			\end{center}
		\end{figure}

\begin{figure}[tb]
			\begin{center}
				\centerline{\includegraphics[width=0.35\columnwidth]{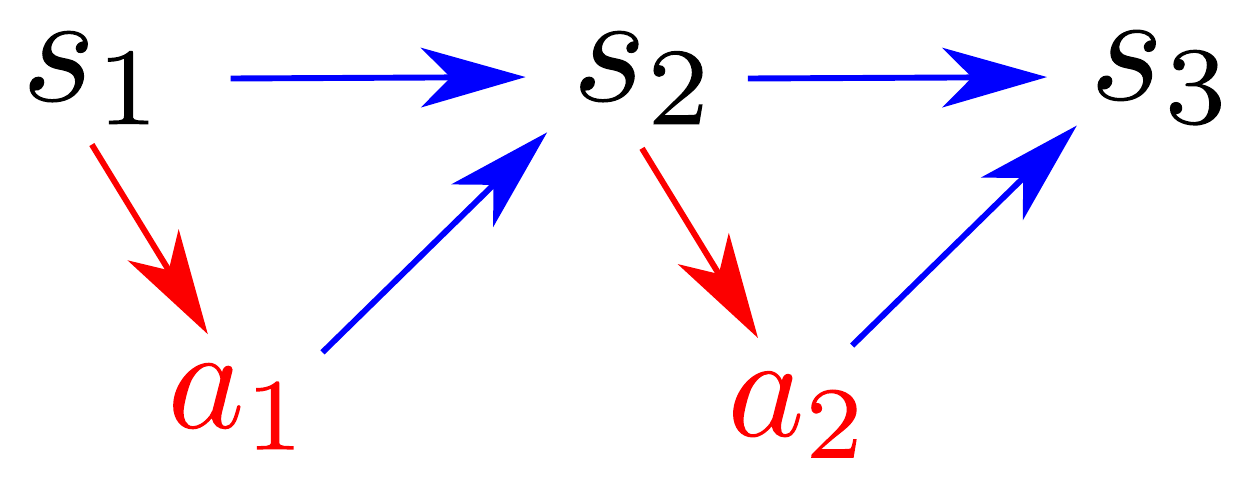}}
				\caption{This is the dependence structure of the states $s$ and actions
					$\bar{a}$ implied by \cref{fig:variables_dependencies_approximation},
				ignoring the light-blue arrows.}
				\label{fig:variables_dependencies_approximation_condensed}
			\end{center}
		\end{figure}

		In the case of low-rate control, we can also define an approximate
		mapping of the real-time system to
		a standard MDP which does not
		require state augmentation. 
		This may in some cases be preferable, because it is more intuitive
		and it does not lead to an observation space of increased
		dimensionality.
		To see how this can be done, consider
		\cref{fig:variables_dependencies_approximation}.
		If we control at a rate reduced by a factor $k$, each control action $a_t$ is
		applied $k$ times and hence affects $k$ observations. For $k$ sufficiently
		large, a reasonable approximation is to ignore the dependence of the last of
		these $k$ observations on $a_t$, see
		\cref{fig:variables_dependencies_approximation}. Doing so leads again to
		standard MDP dependencies, see
		\cref{fig:variables_dependencies_approximation_condensed}.
		We can now define again an according OpenAI Gym environment:
		\begin{example}[Gym Env - Approximation at Reduced
			Rate]\label{para:gym_approximate_reduced} {
				\small
				\setlength{\abovedisplayskip}{6pt}
				\setlength{\belowdisplayskip}{\abovedisplayskip}
				\setlength{\abovedisplayshortskip}{0pt}
				\setlength{\belowdisplayshortskip}{3pt}
				\begin{align*}\texttt{def } & \texttt{step(\ensuremath{a}):}\\
					  & \texttt{for \_ in range(\ensuremath{k\text{):}}}                                           \\
					  & \texttt{\ensuremath{\qquad}\ensuremath{t} = robot.append\_desired\_action(\ensuremath{a})} \\
					  & \texttt{\ensuremath{s'} = robot.get\_observation(\ensuremath{t})}                          \\
					  & \texttt{return \ensuremath{s'}, some\_other\_stuff}                                        
				\end{align*}
			}
		\end{example}
		Here, the time index of the last appended robot action is $t$, and we return
		$s'=y_t$. Hence, all robot actions up to time index $t-1$ have to be executed
		before we can retrieve this observation, and only then the step function will
		return. Nevertheless, the step function will have to be called again before
		robot action $t+1$ can start. This means that the controller will only have one
		robot cycle to compute the next action, despite the lower control rate. 
						
		Summarizing, we can say that this approximation is a good solution if the
		computation of the next action is substantially faster than the control rate.
		Otherwise, the solution in \cref{para:gym_state_augmentation_reduced} is
		preferable, since there the controller may use the entire control cycle to
		compute the next action.
		\subsection{Back-end and Driver}
		To integrate a new robot into our software framework, the only necessary steps
		are to define an \texttt{Action}, an \texttt{Observation}, and a
		\texttt{RobotDriver} class in C++ with the following functions: {\small
			\setlength{\abovedisplayskip}{6pt}
			\setlength{\belowdisplayskip}{\abovedisplayskip}
			\setlength{\abovedisplayshortskip}{0pt}
			\setlength{\belowdisplayshortskip}{3pt}
			\begin{align*}
				  & \texttt{RobotDriver}                                            \\
				  & \texttt{\{}                                                     \\
				  & \qquad\texttt{Observation get\_latest\_observation();}          \\
				  & \qquad\texttt{Action apply\_action(Action \ensuremath{a_{t}});} \\
				  & \texttt{\}}                                                     
			\end{align*}
		} The first function is self-explanatory. The second function takes a desired
		action as input, it may perform safety checks, and then returns the applied
		action.
						
		The back-end takes care of calling the right functions at the right times. It
		will read the desired actions, appended by the user to the timeseries $a$, and
		pass them to the \texttt{RobotDriver}, and it will fill the observation $y$ and
		applied action $a'$ timeseries with the outputs from the \texttt{RobotDriver}.
		For details we refer the interested reader to the documentation in the code.
		\subsubsection{Implementation of the TriFinger Robot}
		\label{sec:trifinger_implementation}
		The implementations of the classes above for the TriFinger robot are:
		\begin{itemize}
			\item \texttt{Observation}: Joint position, velocity and torque of each
			      joint. The camera images are retrieved through a separate
			      interface with the same structure, since they arrive at a different rate.
			\item \texttt{Action}: Torque to be applied at each joint and
			      optionally a joint position along with the gains of a PD
			      controller. The driver will then sum the torque with the
			      feedback from the user-specified position controller.
			\item \texttt{RobotDriver}: Here, the implementation of \texttt{get\_latest\_observation}
			      returns the latest measurements which were received from the motorboards
			      through CAN, see \cite{Grimminger2020-tl} for details. The
			      \texttt{apply\_action} makes sure that the joint velocity and torques do not
			      become too large, since otherwise the robot may break or the motors motors
			      may overheat. It then sends the modified action to the motorboards and
			      returns it.
		\end{itemize}
		\subsection{Safety Checks}
		While the robot hardware is very robust, some additional software safety
		checks are necessary for ensuring that the user code cannot break the
		robot. There are five main checks we perform:
		\begin{itemize}
			\item The backend continuously monitors the timing of received actions
			in a real-time loop. If the expected rate of 0.001s is exceeded
			substantially, the robot is shut down.
			\item There is an additional time-out on the motor board. For instance, in
			case the computer crashes and the motor board does not receive any messages
			for some time it will shut down the motors.
			\item If a joint exceeds a predefined angle, it is brought back into the
			admissible range using a PD controller. This prevents e.g. collision of the
			fingers with the electronics.
			\item We determined the maximum admissible current to prevent overheating of
			the motors, and we ensure that this current is not exceeded by clipping the
			desired torque if necessary.
			\item We simulate joint damping (D-gain) to ensure that the fingers do not
			reach excessive velocities.
		\end{itemize}
		
		Note that we do not prevent collisions (except with the electronics), since the
		hardware design is robust to collisions. In addition, the software design is
		such that users may easily implement their own robot or  even task-specific
		safety checks.
		\subsection{Relation to Robot Operating System (ROS)}
		The core software described above is independent of ROS. We use catkin (which
		can be installed without ROS) for compilation. Further, we use ROS in some of
		the robot-specific packages for peripheral purposes, such as locating other
		packages. Finally, we use Xacro (which is part of ROS) for defining the URDF
		robot model of the TriFinger.
		
		%
		\section{Experiments}
		\begin{figure*}[h]
			\centering
			\includegraphics[width=0.98\textwidth]{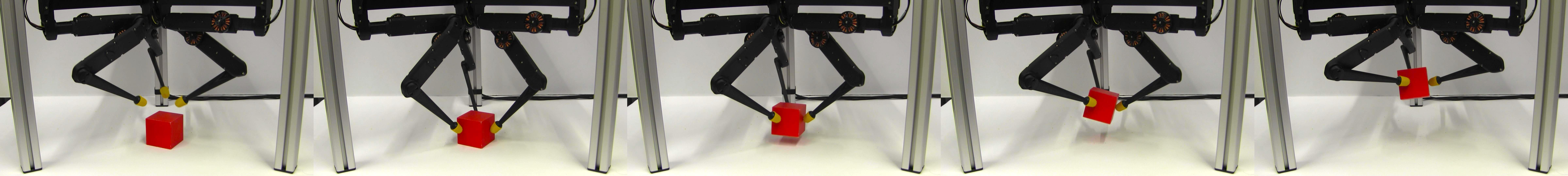}
			\centering
			\caption{TriFinger performing the pickup task.}
			\label{fig:upward_motion}
		\end{figure*}
		The purpose of this section is not to improve the state-of-the art in robotic
		manipulation, but rather to illustrate the capabilities of the proposed hardware
		and software. Each experiment highlights different aspects: 
		\begin{itemize}
			\item \Cref{sec:exp:optimal_control} demonstrates the \SI{1}{kHz} real-time
			torque-control abilities and the ease-of-use of the software interface for
			classical control loops.
			\item \Cref{sec:exp:reinforcement_learning} shows that the backend safety
			features allow for deep reinforcement learning from scratch, without any
			safety checks on the user side. This experiment also shows that the hardware
			is robust against collisions, which will necessarily occur during the
			learning of manipulation tasks. In addition, this use case shows that the
			software interface allows for application of out-of-the-box implementations
			of deep RL methods.
			\item In \cref{sec:exp:throwing}, we perform a throwing experiment to show 
			that the actuators allow for highly-dynamic motions.
			\item In \cref{sec:exp:fine_manipulation} we show through 
			demonstration experiments that the platform is capable of fine-manipulation.
			\item Finally in \cref{sec:exp:endurance} we discuss some experiments assessing 
			the durability of the design.
		\end{itemize}
		Videos of the experiments are available at \cref{google_site}.
		\subsection{Optimal Control}\label{sec:exp:optimal_control} 
\begin{figure}[H]
			\centering
			\includegraphics[width=\linewidth]{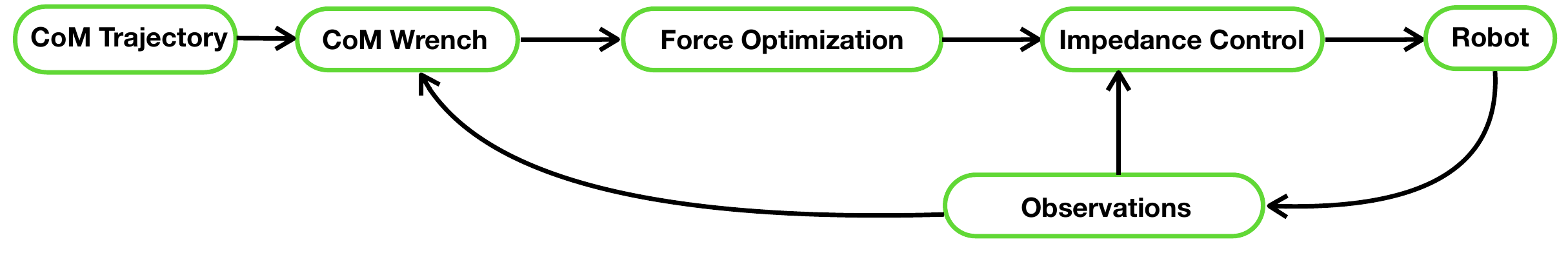}
			\caption{Control loop at 1kHz with force optimization.}
			\label{fig:control_loop_qp}
		\end{figure}

Controlling robot
		interactions with the environment is challenging due to the unilateral nature of
		the contact constraints and the stiff behavior of contact forces. We tackle this
		problem, for picking up and moving a cube, with the control loop depicted in
		\cref{fig:control_loop_qp}.
		\paragraph{Center of Mass Wrench:}
		The first step is to compute the wrench (force and moment) which need to be
		applied to the object to maintain it on the desired trajectory. We do so using a
		simple PD law 
		\begin{align}
			F_{\text{com}} & = P \delta x_{\text{com}} + D \delta \dot{x}_{\text{com}} - m \Vec{g}\\
			M_{\text{com}} & = P \delta q +  D \delta \omega                                       
		\end{align}
		which will compute a force $F_{\text{com}}$ and moment $M_{\text{com}}$
		at the object center-of-mass, to correct for errors in position ($\delta
		x_{\text{com}}$), linear velocity ($\delta \dot{x}_{\text{com}}$),
		orientation ($\delta q$) and angular velocity ($\delta \omega$). $P,D$
		are the controller gains, $\Vec{g}$ is the gravity vector and $m$ is the object mass.
		\paragraph{Force Optimization:} The next question is what forces the finger tips
		must apply to the object in order to achieve the desired wrench calculated
		above. Therefore, we formulate a quadratic program to find optimal distribution
		of contact forces:
		\begin{align}
			F_\text{tip}=\arg\min_{y}\quad & \frac{1}{2}y^{\top}y       \\
			\text{subject to}\quad         & G y\le h                   \\
			                               & A y=\left(\begin{array}{c} 
			F_{\text{com}}\\
			M_{\text{com}}
			\end{array}\right)
		\end{align}
		where the optimization variable $y$ is the stack of contact forces at the three
		finger tips expressed in the
		local frame of the object. The equality constraint ensures that the tip forces
		produce the desired wrench on the object ($A$ transforms tip forces to
		center-of-mass wrench given known contact locations, see e.g.
		\cite{murray2017mathematical}).
		\begin{figure}[H]
			\centering
			\includegraphics[width=.6\linewidth]{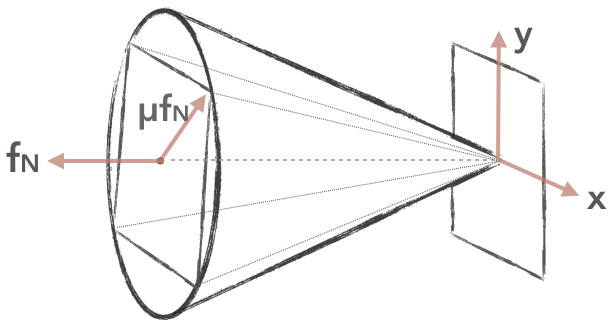}
			\caption{The friction cone ($ \lvert f_p \rvert \leq
				\frac{\mu}{\sqrt{2}} f_n$) can be approximated using linear
				inequalities such that the approximation lies inside the actual cone.
				$\mu$ is the static
				friction coefficient and $f_n, f_p$ are the force components normal
			and parallel to the contact surface, respectively.}
			\label{fig:friction_cone}
		\end{figure}
		The inequality constraint enforces that the tips can push but not pull, and it
		ensures that the object does not slip. It is a linear approximation to the
		friction cone (see \cref{fig:friction_cone} and \cite{murray2017mathematical}
		for details).
		\paragraph{Impedance Control:}
		The final step is to compute the torques $\tau$ to be applied at the robot
		joints to produce the desired tip forces $F_\text{tip}$ computed in the previous
		step. This can easily be achieved via Jacobian ($J$) transpose control. For
		stabilization, we also add a feedback on the fingertip position and velocity
		errors $\delta x_\text{tip}, \delta\dot{x}_\text{tip}$ (the desired tip
		trajectories are obtained given the desired object trajectory and known contact
		locations). This yields a simplified version of the impedance controller
		introduced by \citet{hogan1984impedance}
		\begin{align}
			\tau = J^{\top} \left(F_\text{tip} + P'\delta x_\text{tip} +  D' \delta \dot{x}_\text{tip}\right) 
			\label{eq:finger_impedance}                                                                       
		\end{align}
		where $P'$ and $D'$ are hand-tuned controller gains.
		\subsubsection{Results}
		We apply this methodology to two tasks: Lifting a cube \SI{20}{cm} along a
		vertical line (see \cref{fig:upward_motion}) and sliding that same cube in a
		circle along the table. As can be seen in the videos at \cref{google_site}, the
		forces applied to the object are appropriate for moving it along the desired
		trajectory without slippage. 
		\subsection{Reinforcement Learning}\label{sec:exp:reinforcement_learning}
		\begin{figure}[t!]
			\centering 
			\centering
			\includegraphics[width=0.8\columnwidth]{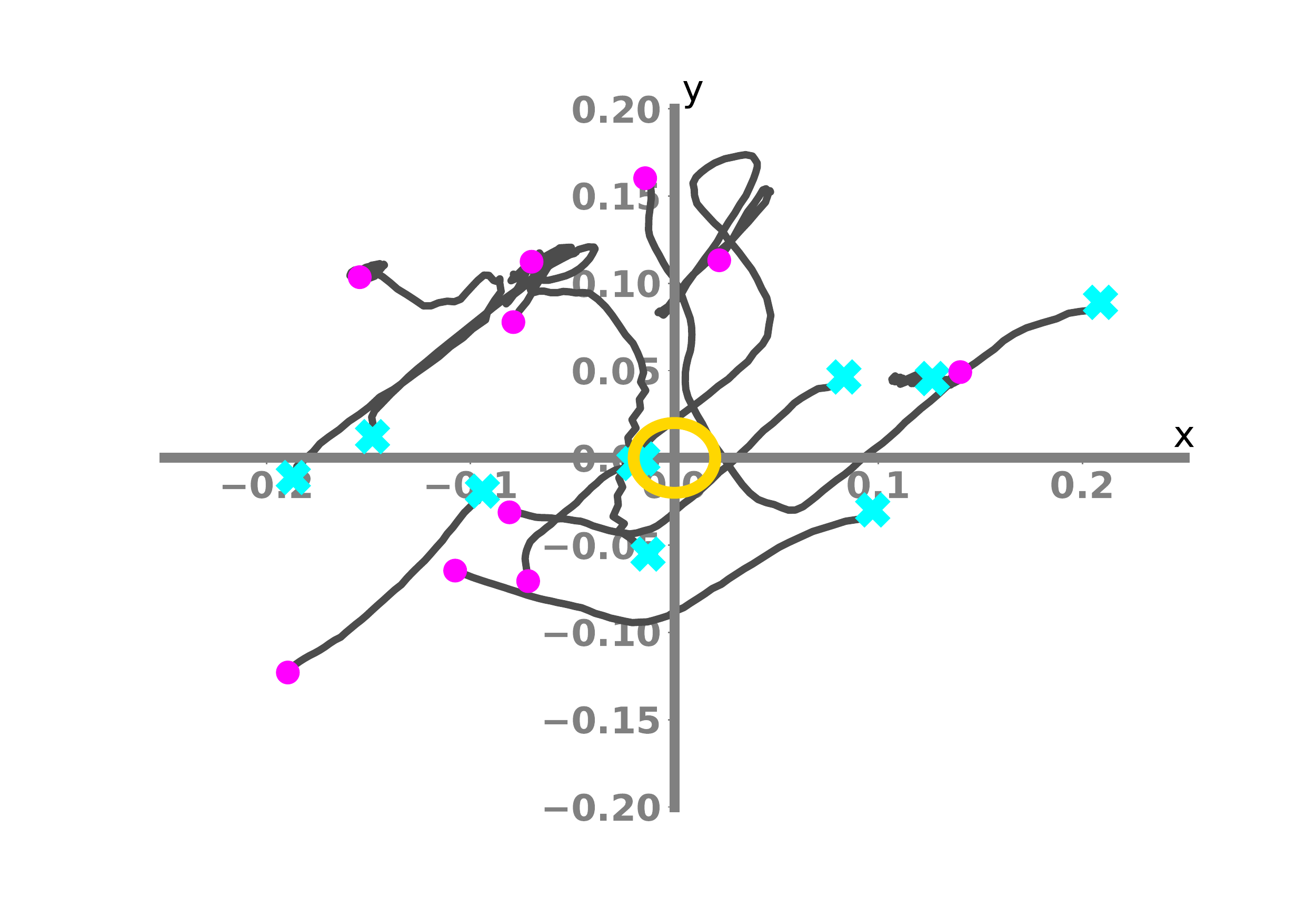}
			\caption{Beginning of training}
			\includegraphics[width=0.8\columnwidth]{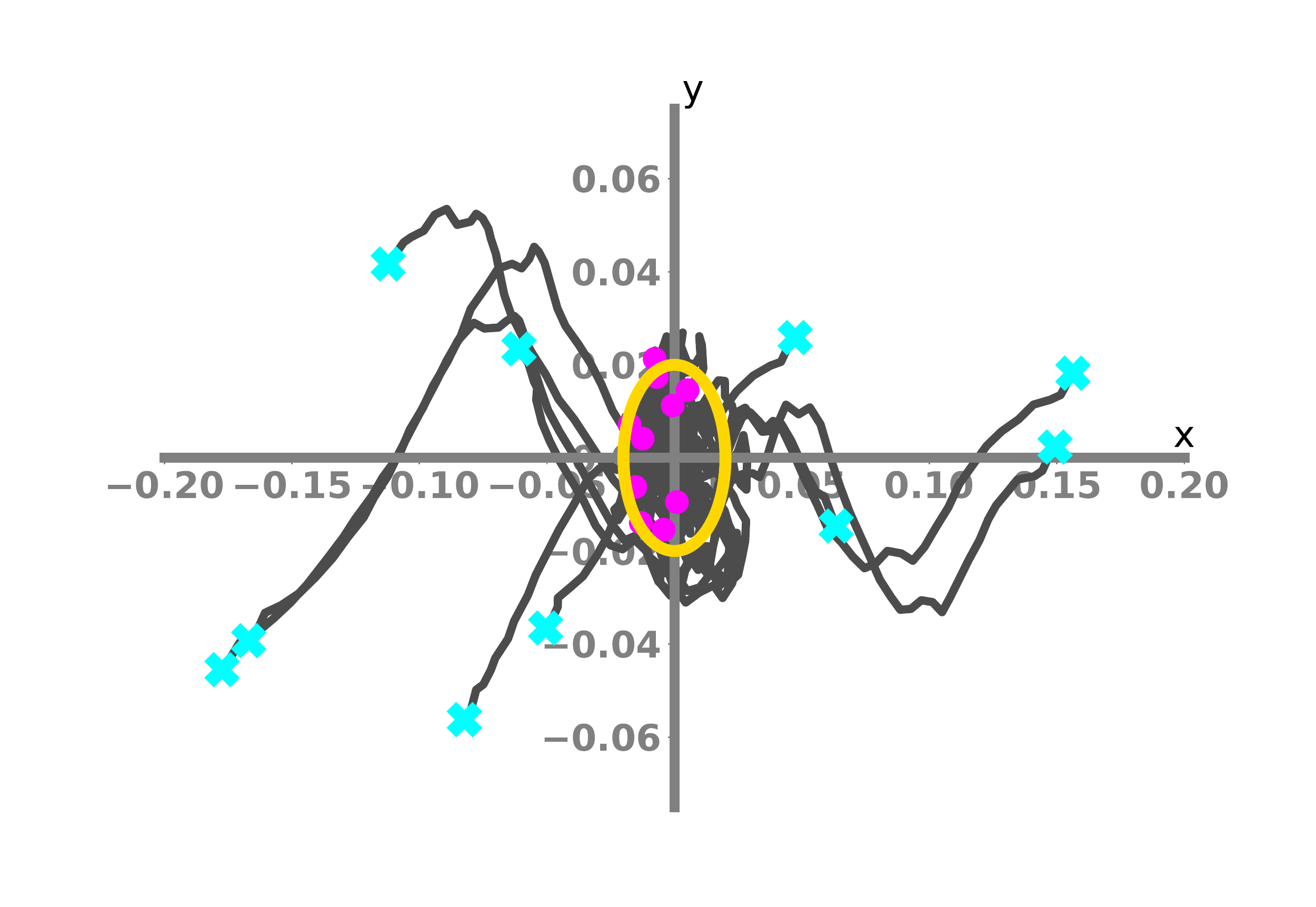}
			\caption{After 200 episodes}
			\centering
			\includegraphics[width=0.8\columnwidth]{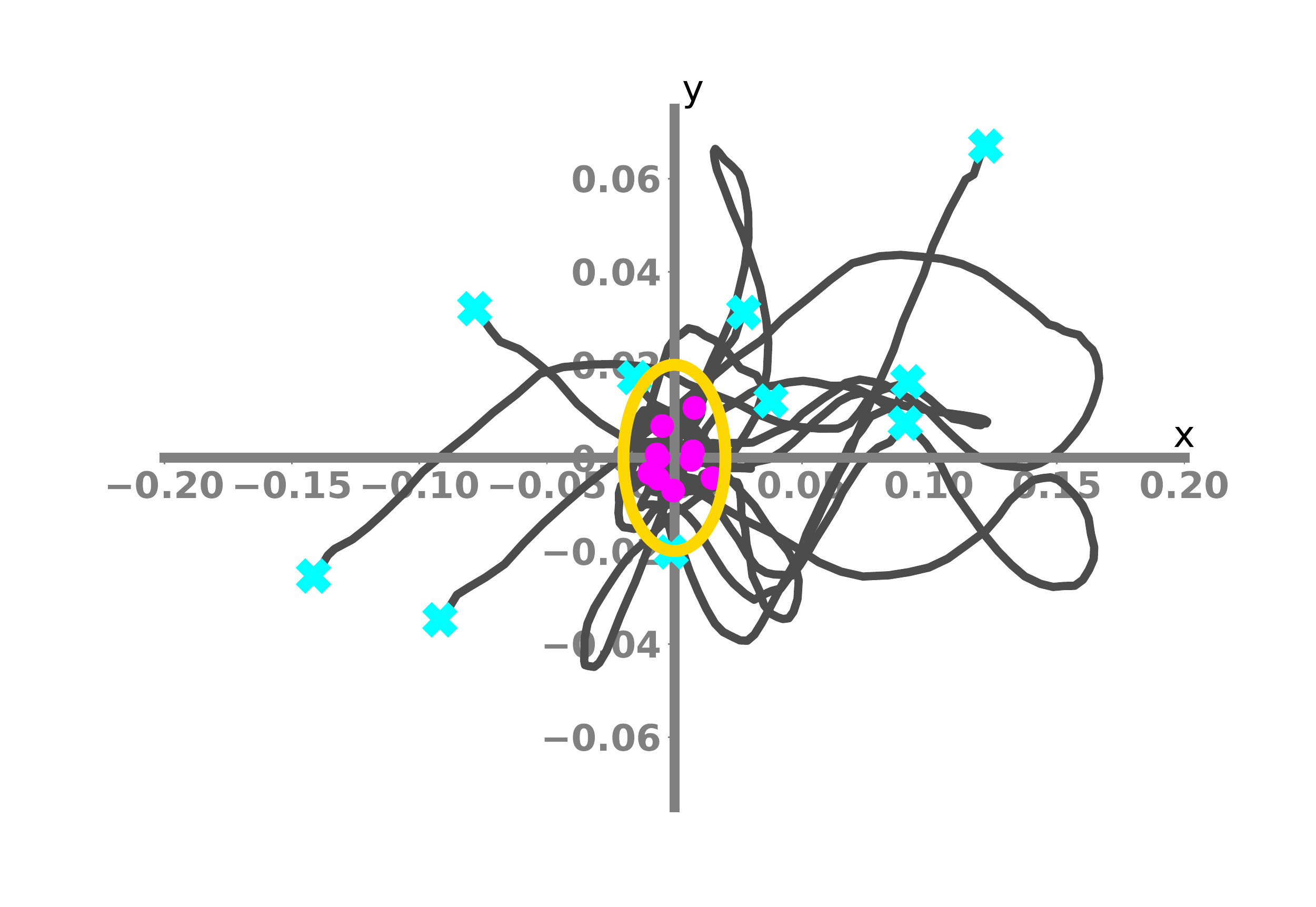}
			\caption{End of training}
			\caption{Trajectories of the finger tip relative to the goal position in the xy-plane. Each trajectory corresponds to one episode (start: cyan cross, end: pink dot). The yellow ellipse marks 2\,cm position error on each axis.} 
			\label{fig:plot_learning_tip_position}
		\end{figure}
		We illustrate the suitability of the platform for real-time reinforcement
		learning by training a DDPG \cite{ddpg} agent from scratch on a reaching task,
		using the DDPG implementation from stable-baselines \cite{stable-baselines}.
						
		In this task, the goal is for each finger-tip to reach a randomly-sampled target
		location as accurately as possible. 
		The episode length is set to 2 seconds. The observation space of the policy
		consists of joint positions, joint velocities, and target positions
		for each finger specified in task space. The action is the desired joint
		configuration of the fingers (the software back-end provides a PD controller, 
		see \cref{sec:trifinger_implementation}).
		The reward at each timestep is the negative Euclidean distance between the
		end-effector and the target. We train the system for 700 episodes, corresponding
		to 23 minutes of execution on the robot, plus a few minutes of computation time
		used by DDPG.
						
		At the beginning of training, the fingers' motions are jerky, and they often
		collide, see the video at \cref{google_site}. As the training progresses, the
		motion becomes smoother and more accurate, see
		\cref{fig:plot_learning_tip_position}. At the end of training, the fingers are
		able to reach the target positions consistently within an error of about 2cm.
		\subsection{Throwing}\label{sec:exp:throwing} To showcase the ability of
		performing highly dynamic tasks, we execute throwing motions recorded through
		kinesthetic teaching (i.e. the motion was demonstrated by guiding the robot
		fingers). The videos at \cref{google_site} show that the TriFinger is able to
		throw light objects several meters. We expect that using appropriate
		controllers, instead of kinesthetic teaching, one could improve considerably on
		these results.
		\balance
		\subsection{Fine Manipulation}\label{sec:exp:fine_manipulation} Finally, to
		illustrate the dexterity of the platform, we perform several fine manipulation
		motions, which, as above, were demonstrated through kinesthetic teaching. As can
		be seen in the videos at \cref{google_site}, the experiments include flipping a
		cube, turning it with one finger while the others hold it, balancing a flat
		cuboid on its side, and picking up a pen and drawing.
		\subsection{Durability Experiments}\label{sec:exp:endurance} 
		We ran durability experiments on a single finger, executing a fast motion in
		free space. In the first experiment a timing belt broke after 79 days of
		continuous operation. This may be partially due to the nonuniform stress put
		on the timing belt by such repetitive motions, which would be less of an
		issue in realistic operation. In the second experiment the shell of the
		center link broke after 72 days of continuous operation, the design has been
		improved since to avoid such breakage.
		
		In addition, we performed some tests on a TriFinger version used in a robot
		competition hosted at our
		institute\footnote{\url{https://real-robot-challenge.com/}}. That version,
		called TriFingerPro, was developed for internal use and is too complex for
		open-sourcing. Nevertheless, it is essentially identical in terms of
		kinematics and actuation, and we would expect its durability to be
		indicative of the open-source version. We executed random motions, including
		collisions between fingers and with external objects, on two of those
		platforms for one week continuously without breakage. 
		
		These results are naturally not statistically significant, and the
		durability most likely depends on the material used for the 3D printing.
		Nevertheless, they are promising and we believe that we can further improve
		durability by fixing weak points in the design as they emerge.
		\section{Conclusion}
		We presented an open-source robotic platform with novel hardware and
		software. We have shown its i) capabilities for dexterous manipulation,
		ii) suitability for deep RL from scratch thanks to the robustness of the
		hardware and safety checks of the software and iii) ease of use for both
		real-time optimal control as well as deep RL. We hope that these
		factors, in combination with the simple and inexpensive hardware design,
		will motivate many researcher to adopt this platform as a shared
		benchmark for real-world dexterous-manipulation. This could lead to a
		more coordinated effort among different labs and the generation of
		orders of magnitude more real-robot data than was possible thus far.
				


		\balance

		\newpage

		\bibliography{references}
		\balance
		
		\bibliographystyle{icml2020}
		\begin{appendices}
			\crefalias{section}{appendix}
			\crefalias{subsection}{appendix}
			\newpage
		
			\section{Comparison with D'Claw}\label{appendix:comparison_d_claw} 
			
			Here we provide a more detailed comparison between the actuator module
			\citep{Grimminger2020-tl} used in the TriFinger and the D'Claw actuators.
			The key specifications of the two robots are given in this table:
			
			\begin{center}
				\begin{tabular}{ l c c }
				  & Gear Ratio & Speed [rpm] \\ 
				  \hline
				 D'Claw & 212.6:1 & 77 \\  
				 TriFinger & 9:1 & 416    
				\end{tabular}
			\end{center}
			
			More details about the TriFinger motor can be found on the site of the
			manufacturer \footnote{\url{https://store-en.tmotor.com/goods.php?id=438}}
			(note that the values above are obtained from the motor specifications and
			the gear ratio of the transmission) and more details for the dynamixel
			module used in D'Claw can be found on the site of Robotis~\footnote{\url{http://www.robotis.us/dynamixel-xm430-w210-r/}}.

			Hence, the maximum speed of the TriFinger joints is 5.4 times faster, which
			allows for more dynamic motions. Further, the gear ratio of the D'Claw is 23.6
			times higher. A higher gear ratio leads to more friction in the transmission
			and a higher motor inertia seen at the joint level (since the rotor has to
			rotate 212.6 times faster than the joint). This leads to more resistance of
			the joint to external forces, which implies large internal forces on the
			transmission and hence increased risk of breakage. 
		\end{appendices}
		
\end{document}